\documentclass[pdflatex,sn-mathphys-num]{sn-jnl}

\usepackage{graphicx}%
\usepackage{multirow}%
\usepackage{amsmath,amssymb,amsfonts}%
\usepackage{amsthm}%
\usepackage{mathrsfs}%
\usepackage[title]{appendix}%
\usepackage{xcolor}%
\usepackage{textcomp}%
\usepackage{manyfoot}%
\usepackage{booktabs}%
\usepackage{algorithmicx}%
\usepackage{listings}%
\usepackage{stmaryrd}
\usepackage{caption}
\usepackage{lineno}
\usepackage{makecell}
\usepackage{adjustbox}

\usepackage[ruled,linesnumbered]{algorithm2e}
\usepackage{amsmath}
\usepackage{adjustbox}

\theoremstyle{thmstyleone}%
\theoremstyle{thmstyletwo}%

\theoremstyle{thmstylethree}%

\begin{document}
\title[Article Title]{Physically consistent and uncertainty-aware learning of spatiotemporal dynamics}

\author[1]{\fnm{Qingsong Xu}}\email{qingsong.xu@tum.de}

\author[1,3]{\fnm{Jonathan L Bamber}}\email{J.Bamber@bristol.ac.uk}


\author[1]{\fnm{Nils Thuerey}}\email{nils.thuerey@tum.de}

\author[1,5]{\fnm{Niklas Boers}}\email{n.boers@tum.de}

\author[3]{\fnm{Paul Bates}}\email{Paul.Bates@bristol.ac.uk}

\author[4]{\fnm{Gustau Camps-Valls} }\email{gustau.camps@uv.es}

\author[1]{\fnm{Yilei Shi} }\email{yilei.shi@tum.de}

\author[1,2]{\fnm{Xiao Xiang Zhu*} }\email{xiaoxiang.zhu@tum.de}

\affil[1]{\orgname{Technical University of Munich}, \postcode{80333} \city{Munich}, \country{Germany}}
\affil[2]{\orgname{Munich Center for Machine Learning},  \postcode{80333} \city{Munich}, \country{Germany}}
\affil[3]{
\orgname{University of Bristol},  \city{Bristol BS8 1SS}, \country{UK}}
\affil[4]{
\orgname{Universitat de Val\`encia},  \city{Val\`encia}, \country{Spain}}
\affil[5]{\orgdiv{Munich Climate Center}, \orgname{Technical University of Munich}, \postcode{80333} \city{Munich}, \country{Germany}}



\abstract{Accurate long-term forecasting of spatiotemporal dynamics remains a fundamental challenge across scientific and engineering domains. Existing machine learning methods often neglect governing physical laws and fail to quantify inherent uncertainties in spatiotemporal predictions. To address these challenges, we introduce a physics-consistent neural operator (PCNO) that enforces physical constraints by projecting surrogate model outputs onto function spaces satisfying predefined laws. A physics-consistent projection layer within PCNO efficiently computes mass and momentum conservation in Fourier space. Building upon deterministic predictions, we further propose a diffusion model-enhanced PCNO (DiffPCNO), which leverages a consistency model to quantify and mitigate uncertainties, thereby improving the accuracy and reliability of forecasts. PCNO and DiffPCNO achieve high-fidelity spatiotemporal predictions while preserving physical consistency and uncertainty across diverse systems and spatial resolutions, ranging from turbulent flow modeling to real-world flood/atmospheric forecasting. Our two-stage framework provides a robust and versatile approach for accurate, physically grounded, and uncertainty-aware spatiotemporal forecasting.}

\maketitle

\section{Introduction}\label{sec1}

\label{Introduction}
Prediction and modeling of spatiotemporal dynamics governed by partial differential equations (PDEs) remain a central challenge in scientific and engineering research~\cite{rao2023encoding, zappala2024learning}, with applications spanning Earth system modeling~\cite{bodnar2025foundation, irrgang2021towards}, fluid dynamics~\cite{vinuesa2023transformative}, and geophysics~\cite{yu2021deep}. Traditional fluid dynamics simulators often rely on manually crafted simplifications and incur substantial computational costs, as exemplified by the Navier–Stokes equations for fluid flow and the shallow water equations for flood and atmospheric modeling~\cite{kochkov2021machine,xu2025physics}.

Machine learning-based approaches have recently emerged as promising alternatives, demonstrating notable efficiency in capturing complex dynamics. Physics-informed neural networks (PINNs)~\cite{raissi2019physics,karniadakis2021physics} employ continuous learning paradigms, using neural networks to approximate solutions of physical systems. However, PINNs require explicit knowledge of the governing PDEs and cannot inherently encode prior physical information into the model~\cite{fuks2020limitations}.
Spatiotemporal discrete learning methods, such as Fourier neural operators (FNOs)~\cite{li2020fourier}, deep operator networks~\cite{lu2021learning}, and Laplace neural operators (LNOs)~\cite{cao2024laplace}, learn mappings between function spaces and have shown success in simulating diverse PDE systems without retraining for new conditions. These methods can explicitly enforce physical constraints and partially encoded PDE structures within the learning process. Furthermore, frequency domain-based discrete learning approaches, such as FNOs, offer superior advantages in effective feature representation and resolution invariance through the fast Fourier transform.

Despite significant advancements in discrete learning methods, these models remain largely data-driven and often fail to incorporate the intrinsic physical laws embedded within the data. Consequently, their performance strongly depends on the quantity and diversity of available training data and they struggle to maintain physical consistency over long-term, large-scale spatiotemporal predictions, such as climate projections or flood forecasting, where deviations can accumulate and significantly degrade rollout accuracy~\cite{xu2024large}. 

Several neural PDE solvers have sought to embed intrinsic physical properties within network architectures to improve the efficiency of learning underlying physical phenomena. Notable strategies include the design of neural networks that incorporate symmetries, such as group-equivariant FNOs (G-FNOs)~\cite{helwig2023group}, symmetry-enforcing frameworks inspired by Noether’s theorem~\cite{muller2023exact,xu2024physics}, physics-encoded recurrent convolutional neural network that preserve structural constraints~\cite{rao2023encoding}, and stabilization of neural differential equations \cite{white2023stabilized}. Additional strategies enforce mass conservation through architectural or algorithmic modifications, such as implicit differentiation~\cite{smith2022physics} and mass conservation law-encoded FNO (ClawFNO)~\cite{liu2024harnessing} by constructing divergence-free conditions through antisymmetry.

Despite these efforts, existing studies have not systematically addressed the simultaneous conservation of both mass and momentum at the architectural level for discrete learning models. Furthermore, incorporating uncertainty quantification into predictive frameworks is essential for assessing the reliability of forecasts~\cite{scher2018predicting,hess2025fast}. Nevertheless, most existing neural-PDE solvers formulate spatiotemporal forecasting as a purely deterministic task, thereby failing to account for inherent uncertainties in predicted physical processes. The evolution of chaotic and noise dynamical systems as typically the case in Earth sciences is intrinsically uncertain, where even minor local errors can accumulate and propagate across scales, producing substantial deviations from true dynamics. Without reliable uncertainty estimates, a forecast is of limited value. 

\begin{figure}[htp!]
	\centering
	{\includegraphics[width = 1.01\textwidth]{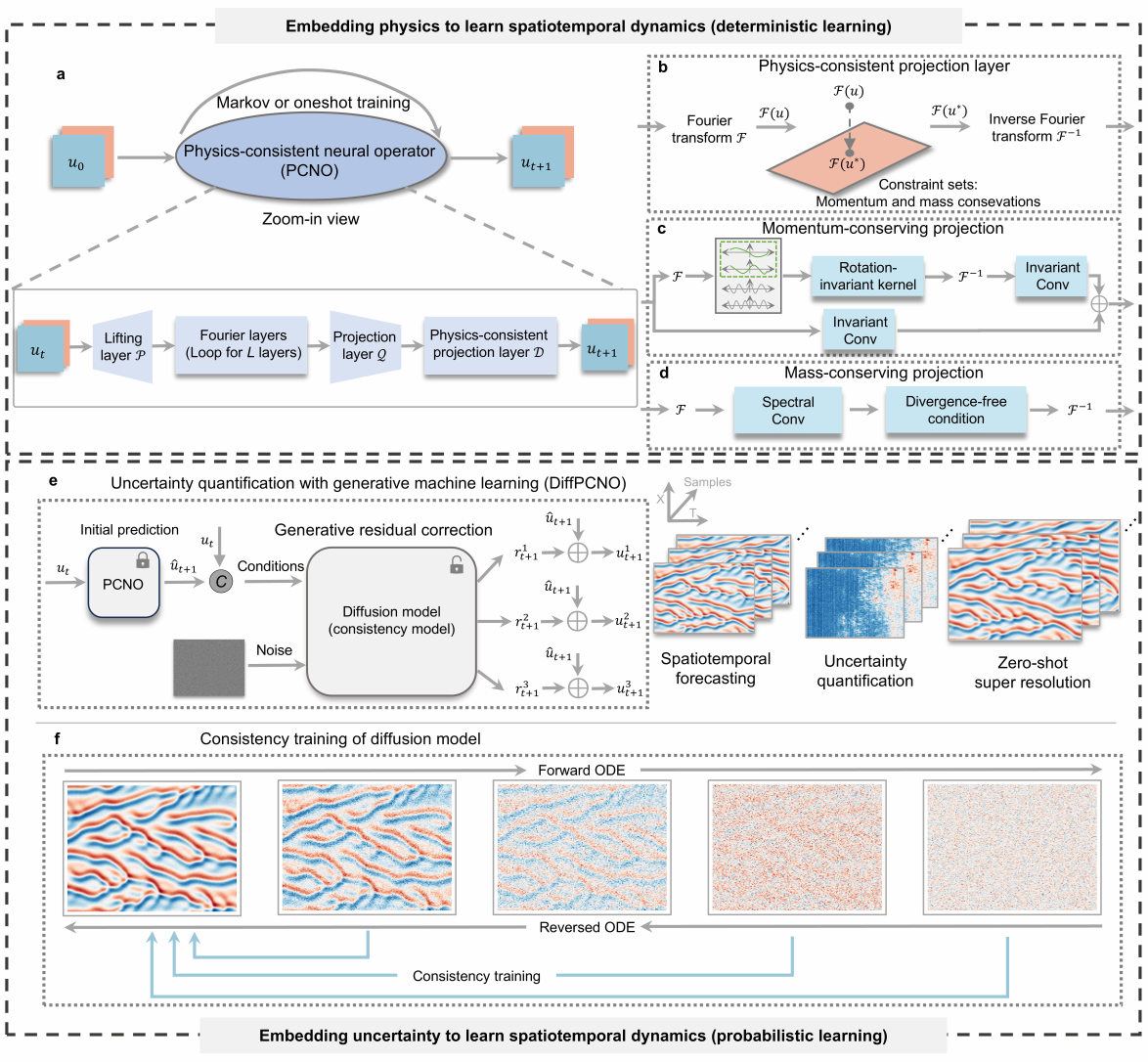}}
	\caption{Schematic illustration of the proposed PCNO and DiffPCNO frameworks. \textbf{a}, Embedding physics through PCNO. PCNO enforces physical consistency by projecting surrogate outputs onto function spaces that satisfy predefined physical laws via a physics-consistent projection layer. \textbf{b}, Schematic of the physics-consistent projection layer, which enforces mass and momentum conservation in spatiotemporal dynamics. \textbf{c}, Momentum-conserving projection with invariance via a rotation-invariant kernel in Fourier space. \textbf{d}, Mass-conserving projection by embedding the divergence-free condition in Fourier space. \textbf{e}, Embedding uncertainty through DiffPCNO corrects prediction residuals via a generative residual correction mechanism based on a consistency model, conditioned on the deterministic prediction $\mathbf{\hat{u}}_{t+1}$ from PCNO and the current state $\mathbf{u}_t$. It aims to capture the residual distribution $\mathbf{r}_{t+1}=\mathbf{y}-\mathbf{\hat{u}}_{t+1}$, where 
$\mathbf{y}$ denotes the ground-truth solution. It enables applications such as spatiotemporal forecasting, uncertainty quantification, and zero-shot super-resolution (downscaling). For instance, DiffPCNO can rapidly estimate uncertainty by generating multiple samples of the dynamical process. \textbf{f}, Consistency models are a class of diffusion-based generative models that generate high-quality samples in a single step while retaining the flexibility for multi-step sampling to balance computational cost and fidelity. For DiffPCNO, an improved consistency model based on the probability flow ordinary differential equation (ODE) is employed and trained using consistency training.}
	\label{fig:1}
\end{figure}

We address these limitations by proposing a physics-consistent neural operator (PCNO) (Fig.~\ref{fig:1}a) for embedding physics to learn spatiotemporal dynamics. PCNO enforces conservation of momentum and mass by projecting surrogate model outputs onto function spaces that satisfy physical constraints in a transformed domain. A physics-consistent projection layer is developed to efficiently impose these constraints through transformations in Fourier space.
Specifically, leveraging Noether’s theorem, the projection layer preserves linear and angular momentum by enforcing translation and rotation invariance in Fourier space. Mass conservation is ensured by imposing divergence-free conditions in Fourier space. PCNO is compatible with a wide range of surrogate models and integrates seamlessly into existing neural operators. Here, FNO serves as a representative backbone. PCNO achieves high-accuracy spatiotemporal predictions while maintaining physical consistency across turbulent flows, flood forecasting, and atmospheric dynamics. 

To address uncertainty in spatiotemporal forecasting, we propose a consistency model-based probabilistic learning framework, the diffusion model-enhanced PCNO (DiffPCNO) (Fig.~\ref{fig:1}e), which builds upon the deterministic predictions of PCNO. Consistency models~\cite{song2023consistency} generate high-quality samples in a single step while retaining the flexibility of multi-step sampling to trade computational cost for fidelity. Unlike generative adversarial networks~\cite{goodfellow2014generative}, they avoid adversarial training, and unlike stochastic differential equation-based diffusion models~\cite{ho2020denoising,song2021denoising}, they do not require costly iterative denoising, which is impractical for large-scale systems such as floods or climate dynamics. Here, we enhance efficiency by integrating an improved consistency model~\cite{song2024improved} with consistency training into PCNO. Specifically, DiffPCNO refines prediction residuals through a generative residual correction mechanism, enhancing long-term predictive accuracy. Experiments demonstrate that the spatial distribution of uncertainty generated by DiffPCNO closely aligns with the relative error distribution, reflecting error propagation in spatiotemporal dynamics. This alignment indicates that embedding uncertainty into the prediction process provides a reliable measure of predictive confidence, which is particularly valuable for real-world applications, such as flood forecasting. Furthermore, we introduce PCNO-Refiner, which applies a consistency model-based refinement to deterministic predictions of PCNO. PCNO-Refiner produces localized uncertainty distributions aligned with its
predictions, but remains less effective than DiffPCNO in correcting substantial deviations from the ground truth.
\section{Results}\label{sec2}
In this section, we propose a two-stage learning paradigm (Fig.~\ref{fig:1}) that integrates deterministic learning with PCNO, which enforces physical constraints within the output space of any surrogate model to learn spatiotemporal dynamics, and probabilistic learning with DiffPCNO, which employs a consistency model to incorporate uncertainty in spatiotemporal dynamics through a generative residual correction mechanism. 
The efficacy of this framework is demonstrated across four representative dynamical systems: the Kuramoto–Sivashinsky dynamics, Kolmogorov turbulent flow, real-world flood forecasting, and atmospheric modeling.
\subsection{Kuramoto–Sivashinsky dynamics}
\begin{figure}[htp!]
	\centering
	{\includegraphics[width = 1.01\textwidth]{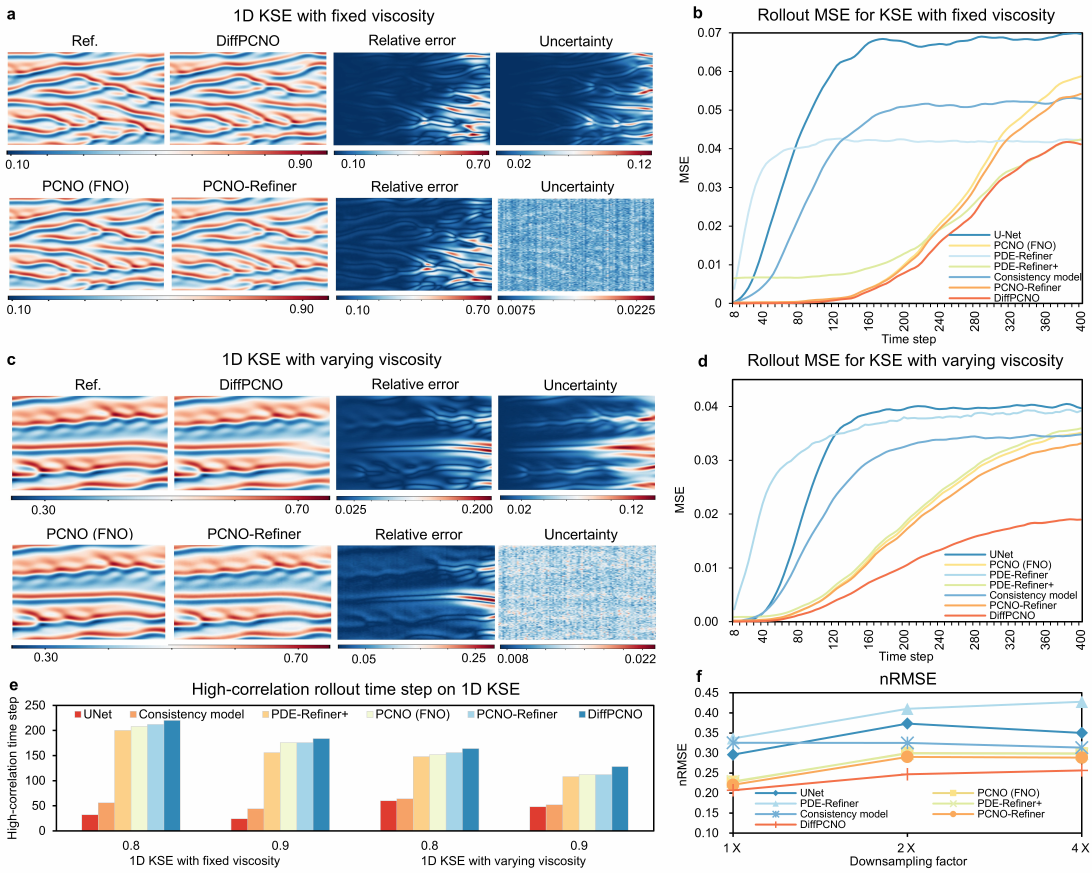}}
	\caption{Results of the one-dimensional Kuramoto-Sivashinsky dynamics. \textbf{a,} Visualization results and uncertainty quantification of DiffPCNO and PCNO-Refiner on the 1D KSE with fixed viscosity $\nu=1$. The vertical axis of the visualization represents 256 spatial points, while the horizontal axis corresponds to the chaotic spatiotemporal evolution over 400 testing time steps. Uncertainty is generated via a stochastic recursive process with 50 sampled trajectories per test case, from which empirical standard deviations are computed (see Methods for detailed information). \textbf{b,} Rollout MSE  of PCNO, DiffPCNO, PCNO-Refiner, and baseline methods for KSE with fixed viscosity over 400 testing time steps. \textbf{c,} Visualization results and uncertainty quantification of DiffPCNO and PCNO-Refiner on the 1D KSE with varying viscosity $\nu$ sampled uniformly between 0.5 and 1.5. \textbf{d,} Rollout MSE  of PCNO, DiffPCNO, PCNO-Refiner, and baseline methods for KSE with varying viscosity over 400 testing time steps. \textbf{e,} Time steps where the average correlation drops below 0.9 and 0.8, indicating the temporal horizon of reliable predictions. \textbf{f,} Super-resolution performance evaluation of PCNO, DiffPCNO, PCNO-Refiner, and baseline methods, on the KSE with varying viscosity, with models trained on a $64 \times 64$ spatial grid and tested directly on downscaled grids at $1\times$, $2\times$, and $4\times$ resolutions.}
	\label{fig:2}
\end{figure}
We first consider the one-dimensional Kuramoto-Sivashinsky equation (KSE), a fourth-order nonlinear PDE exhibiting rich dynamical features and intrinsic chaotic behavior~\cite{li2024learning,lippe2023pde},
\begin{equation}
\partial_t u + u \, \partial_x u + \partial_x^2 u + \nu \, \partial_x^4 u = 0,
\end{equation}
where $\nu$ denotes the viscosity parameter. 
 Our goal is to obtain solutions $u(x, t)$ for all $x$ and $t$ within a spatial domain $[0, L]$ subject to periodic boundary conditions $u(0, t)=u(L, t)$ and initial condition $u(x, 0)=u_0(x)$. Data are generated on a 256-point spatial grid, with time steps $\Delta t$ drawn uniformly between 0.18 and 0.22 seconds~\cite{brandstetter2022lie}. To develop a surrogate model capable of maintaining high-fidelity predictions over extended temporal horizons, training trajectories span $140 \Delta t$, while testing extends to $400 \Delta t$. Due to the chaotic nature of the KSE, variations in viscosity induce markedly different spatiotemporal evolutions. 
 We therefore consider two scenarios: the KSE with fixed viscosity $\nu=1$, and the parameter-dependent KSE with varying viscosity $\nu$ sampled uniformly between 0.5 and 1.5 (see Supplementary Notes 1 for details).
 

 The KSE does not satisfy either momentum or mass conservation, allowing PCNO to directly employ FNO as its surrogate model. We evaluate the performance of PCNO combined with a generative consistency model for long-term prediction. Overall, DiffPCNO consistently outperforms other end-to-end baselines (FNO, U-Net, PDE-Refiner using a denoising diffusion probabilistic model (DDPM)-based refinement, PDE-Refiner+ integrating PCNO predictions, and the standalone consistency model) (see details in Methods). Rollout mean squared error (MSE) (see the definition in Methods) for both fixed viscosity (Fig.~\ref{fig:2}b) and varying viscosity (Fig.~\ref{fig:2}d) demonstrates that prediction accuracy progressively declines over extended horizons. Notably, U-Net, PDE-Refiner, and consistency model exhibit rapid error accumulation over 400 steps, emphasizing the intrinsic difficulty of forecasting long-term KS dynamics. In contrast, the diffusion model-enhanced PCNO (DiffPCNO and PCNO-Refiner) maintains high accuracy and stability over long rollouts. We further perform sensitivity analyses on model inputs, architectures, and the integration methods of the diffusion model. Incorporating deterministic predictions from PCNO as conditional inputs significantly improved the generative model’s stability and accuracy, as evidenced by comparisons between PDE-Refiner and PDE-Refiner+, and between the consistency model and PCNO-Refiner. Regarding the architecture, compared to the DDPM-based PDE-Refiner+, the consistency model-based PCNO-Refiner achieves higher generation (lower MSE) fidelity through consistency training. Furthermore, by comparing PCNO-Refiner and DiffPCNO, it is evident that DiffPCNO, which corrects residuals through probabilistic learning, attains higher accuracy. This improvement arises because PCNO-Refiner corrects deterministic predictions from PCNO, which are difficult to adjust when large errors occur. By contrast, DiffPCNO refines the residuals of deterministic predictions, which are more structured and less correlated, thereby promoting the learning of the generative model. These distinct integration methods of the diffusion model also result in different uncertainty results. 

Uncertainty quantification is shown in Fig.~\ref{fig:2}a and Fig.~\ref{fig:2}c for fixed and varying viscosity, respectively.  For DiffPCNO, regions of high relative error exhibit correspondingly higher uncertainties, and uncertainties generally grow with prediction horizon. PCNO-Refiner shows a more localized uncertainty distribution aligned with its prediction results. Further comparison of the uncertainty estimates indicates that PCNO-Refiner exhibits lower uncertainty values than DiffPCNO in regions with large prediction errors. This suggests that, while PCNO-Refiner improves predictions relative to PCNO, it has limited ability to correct results that deviate substantially from the ground truth. In contrast, DiffPCNO employs a generative residual correction mechanism to adjust such deviations, enhancing long-term prediction accuracy. However, DiffPCNO inherently produces high uncertainty values in regions characterized by substantial residuals. Such increased uncertainty reflects meaningful predictive confidence and can be advantageous in practical applications such as climate prediction.

To evaluate the ability of different models to capture the overall trend and fluctuation patterns of the time series, we perform autoregressive rollouts on the test set and compute the Pearson correlation between the predicted and true values (Fig.~\ref{fig:2}e). 
Across different experiments and correlation thresholds, DiffPCNO consistently maintains reliable and stable long-term dynamics of KSE compared with other baseline models. For instance, in the experiment with fixed viscosity, DiffPCNO sustains a high-correlation rollout (correlation $>$ 0.8) up to $220 \Delta t$. 
Furthermore, PCNO-Refiner exhibits enhanced temporal stability relative to PCNO(FNO), particularly when the correlation is below 0.8. 
In addition, our approach demonstrates exceptional zero-shot super-resolution.
Compared with the robust zero-shot super-resolution baseline (FNO), PCNO-Refiner and DiffPCNO (Fig.~\ref{fig:2}f) achieve superior performance in terms of the normalized relative mean squared error (nRMSE) (see Methods for definition).

 
\subsection{Kolmogorov turbulent flow}
\begin{figure}[htp!]
	\centering
	{\includegraphics[width = 1.01\textwidth]{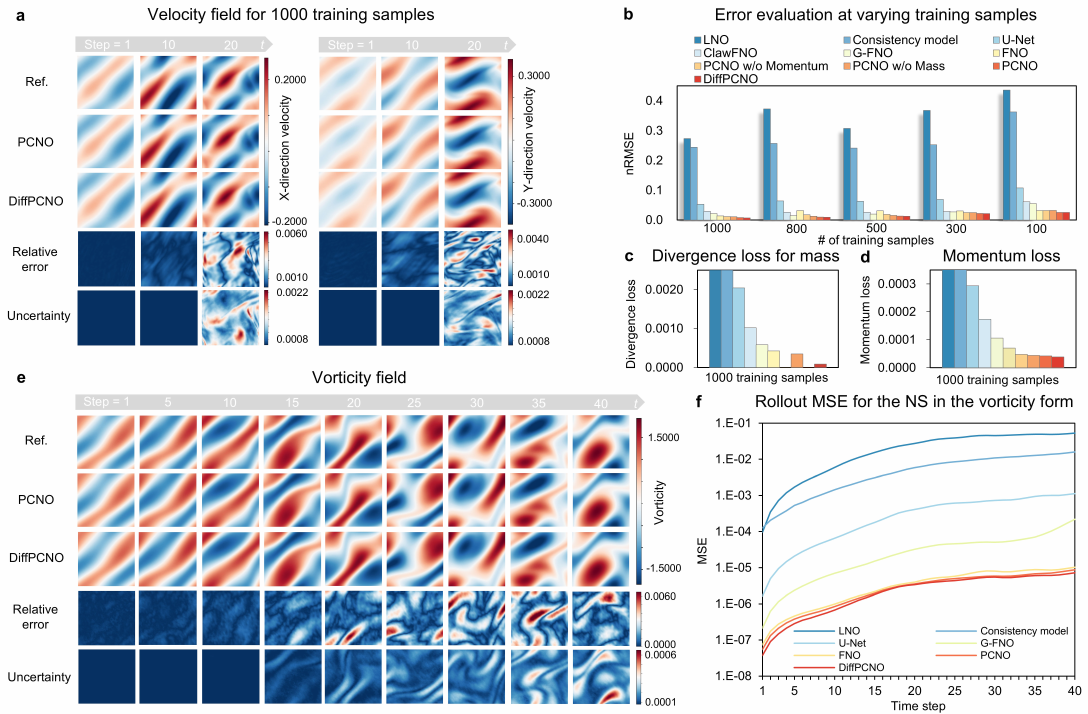}}
	\caption{Results of the two-dimensional Kolmogorov turbulent flow. \textbf{a,} Visualization and uncertainty quantification on the NS equations in velocity form, using datasets comprising 1000 training trajectories. \textbf{b,} Effect of training sample size (100, 300, 500, 800, and 1000 training trajectories) on model performance for the NS equations in velocity form. \textbf{c,} Mass conservation is quantified using divergence loss for the NS equations in velocity form. \textbf{d,} Momentum conservation is quantified using momentum loss for the NS equations in velocity form. \textbf{e,} Visualization and uncertainty quantification on the NS equations in the vorticity form. \textbf{f,} Rollout MSE  of PCNO, DiffPCNO, and baseline methods for the NS equations in vorticity form over 40 testing time steps. The $y$-axis is plotted on a logarithmic scale of the MSE. Across both the NS datasets in velocity form (\textbf{b}) and vorticity form (\textbf{f}), the PCNO architecture consistently improves prediction accuracy. The incorporation of the consistency model via a generative residual correction mechanism into PCNO (DiffPCNO) results in further performance enhancement in both settings. As illustrated in panels \textbf{a} and \textbf{e}, PCNO and DiffPCNO accurately capture the temporal evolution of velocity and vorticity fields across different time scales.}
	\label{fig:3}
\end{figure}
To further evaluate the capabilities of PCNO and DiffPCNO in modeling complex dynamical systems, we consider a two-dimensional spatiotemporal system that conserves momentum and mass, that is, Kolmogorov turbulent flow. Governed by the incompressible Navier–Stokes (NS) equations, the dynamics of Kolmogorov flow are described by,
\begin{equation}
\nabla \cdot \mathbf{u}=0, \quad \partial_t w+\mathbf{u} \cdot \nabla w=\nu \Delta w+\mathbf{f},
\end{equation}
where $\mathbf{u} \in \mathbb{R}^2$ is the velocity field; $w=\nabla \times u$ is the vorticity; $\nu \in \mathbb{R}_{+}$is the viscosity coefficient. $f \in \mathbb{R}$ represents the external forcing term. The system is defined over $x \in(0,1)^2$ and $t=1,2, \ldots, T$. 
We consider the NS equations in velocity and vorticity forms to evaluate the physics-consistent projection layer in PCNO. The model predicts $T=20$ timesteps for the velocity form and $T=40$  timesteps for the velocity form, in both cases conditioned on the first $T_{in}=10$ timesteps (see Supplementary Notes 2 for details).

For the NS in the velocity form, we systematically examine the effect of varying the number of training samples (Fig.~\ref{fig:3}b). PCNO and DiffPCNO achieve consistently lower nRMSE than all end-to-end baselines, including LNO, consistency model, U-Net, vanilla FNO without physics embedding, ClawFNO with mass conservation, group equivariant FNO (G-FNO) with symmetry, even under data-limited regimes. Specifically, comparing PCNO with variants lacking either momentum conservation (PCNO w/o momentum) or mass conservation (PCNO w/o mass) reveals that enforcing both constraints consistently improves accuracy across all training sample sizes. We further evaluate different physical embedding strategies. For mass conservation, we compare PCNO w/o momentum to ClawFNO, which implements a divergence-free constraint using an antisymmetric matrix in the FNO output layer. Our divergence-free approach, implemented through Fourier transforms in the output space, proves more effective, particularly for small training sets (e.g., 100 samples), as it better preserves the form of the continuity equation while the spectral convolution within the mass-conserving projection layer enhances its representational capacity. For momentum conservation, we compare embedding translational and rotational symmetries within the Fourier layer (G-FNO) to our PCNO w/o mass, which enforces momentum conservation directly in the output space. Embedding physical constraints in the output space consistently yields superior performance, particularly for limited datasets. This is because embedding physical constraints in the output space achieves a better balance between physical priors and network expressiveness than embedding them within the network architecture.
The above results collectively demonstrate that embedding physical constraints, including momentum and mass conservations, into the neural network output space enhances the accuracy and robustness of spatiotemporal predictions, particularly in small-data regimes.

Mass and momentum conservations are further quantified using divergence loss and momentum loss, respectively (definitions in Methods). For mass conservation (Fig.~\ref{fig:3}c),
models with mass-conserving projection layers (PCNO and PCNO w/o momentum) achieve near-zero divergence loss, confirming effective mass preservation. DiffPCNO exhibits a slight increase in divergence loss, which is attributable to the diffusion model-based refinement process, as it does not explicitly enforce mass conservation during consistency training. For momentum conservation (Fig.~\ref{fig:3}d), models incorporating the momentum-conserving layer (PCNO and PCNO w/o mass) reduce momentum loss to near-zero levels. DiffPCNO does not compromise this property, as consistency training accelerates convergence toward the true momentum field.

For the long-term NS dataset in the vorticity form, as the output is a univariate vorticity field, the physics-consistent projection layer in PCNO considers only momentum conservation. Rollout MSE of different models (Fig.~\ref{fig:3}f) increases progressively over time, with PCNO consistently exhibiting lower MSE than other baselines, including LNO, consistency model, U-Net, G-FNO, and FNO. 
Moreover, DiffPCNO, enhanced with the diffusion model, further improves the accuracy of PCNO. A comparison between DiffPCNO and the consistency model reveals that incorporating deterministic predictions from PCNO as inputs to the consistency training substantially enhances spatiotemporal forecasting for two-dimensional dynamics. 

We further perform uncertainty quantification of DiffPCNO for spatiotemporal dynamics on the NS dataset in both velocity and vorticity forms (Figs.~\ref{fig:3}a and e). The spatial distribution of uncertainty closely aligns with the corresponding relative errors, and uncertainty values increase over time. These observations demonstrate that embedding uncertainty within the spatiotemporal predictions effectively reflects the error distribution, providing a robust measure of predictive confidence for practical applications, such as two-dimensional flood forecasting.
\subsection{Real-world flood inundation forecasting}
\begin{figure}[htp!]
	\centering
	{\includegraphics[width = 1.0\textwidth]{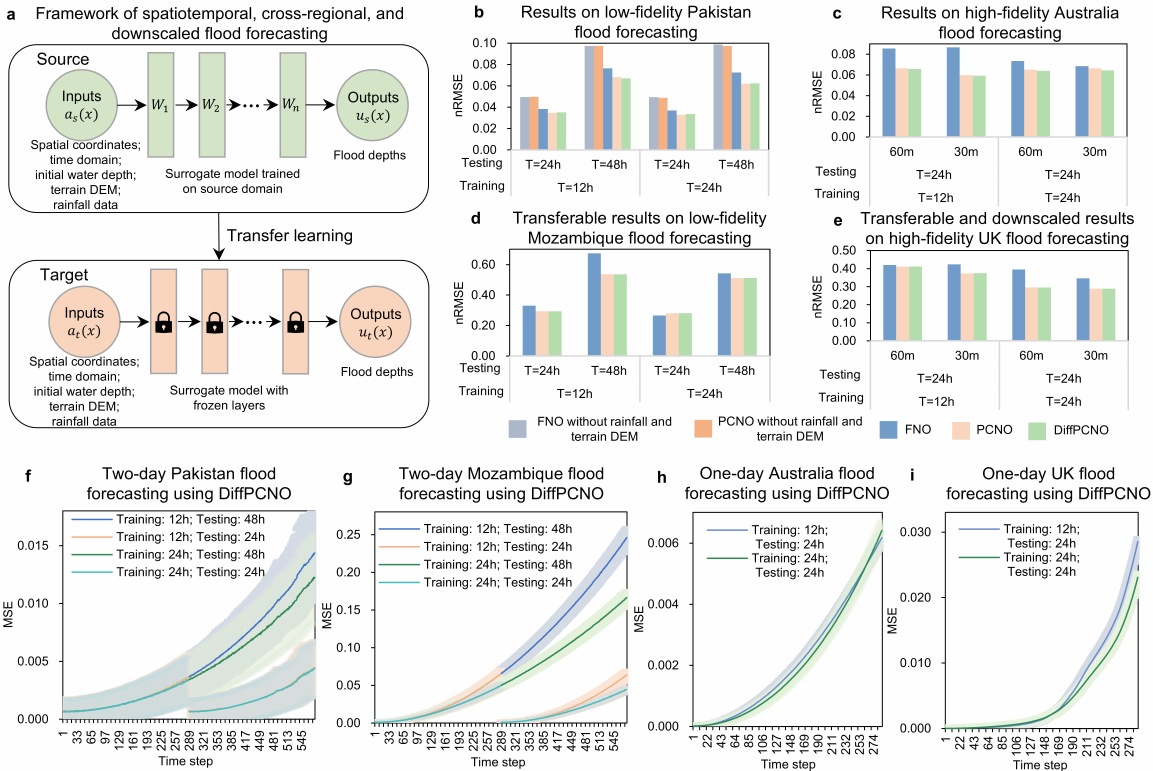}}
	\caption{Results of real-world flood inundation forecasting. 
\textbf{a,} Schematic of spatiotemporal, cross-regional, and downscaled flood forecasting. 
The surrogate models are formulated using input data comprising 2D spatial coordinates, temporal information, initial water depth, DEM, and rainfall forcing over the prediction horizon. 
After training, the surrogate model layers are frozen (locked) and directly deployed for flood forecasting. \textbf{b,} Relative error in low-fidelity Pakistan flood forecasting (480 m spatial, 5 min temporal resolution) across different training and testing horizons: training samples of 144 time steps (12 hours) and testing at 288 (24 hours) or 576 time steps (48 hours); training samples of 288 time steps (24 hours) and testing at 288 or 576 time steps. 
The dataset spans 14 days, with the final 2 days for testing. 
\textbf{c,} Relative error in high-fidelity Australia flood forecasting (60 m training resolution, 30 m downscaling resolution, 5 min temporal resolution) across different training and testing horizons: training samples of 144 time steps (12 hours) and testing at 288 (24 hours) or 576  steps (48 hours); training samples of 288 time steps (24 hours) and testing at 288 or 576  steps. 
The dataset spans 10 days, with the final day for testing. 
\textbf{d,} Relative error of transferable Mozambique flood forecasting (480 m spatial, 5 min temporal resolution). 
The Mozambique flood dataset spans 4 days for testing.  
\textbf{e,} Relative error of transferable and downscaled UK flood forecasting (60 m training resolution, 30 m downscaling resolution, 5 min temporal resolution).  
The UK flood dataset spans 3 days for testing. 
 \textbf{f–i} present rollout MSE and uncertainty for two-day ($T=576$) Pakistan and Mozambique flood forecasts and one-day ($T=288$) Australia and UK flood forecasts using DiffPCNO.}
	\label{fig:4}
\end{figure}
Flooding is a frequent and widespread natural hazard, causing substantial human and economic losses worldwide each year~\cite{tellman2021satellite,rentschler2023global}. This highlights the urgent need for robust flood warning and management systems, particularly in data-sparse watersheds~\cite{nearing2024global}. Accurate flood inundation forecasts are traditionally obtained from two-dimensional hydrodynamic models solving the shallow water equations (SWE) (see Eq. 1 in Supplementary Notes 3)~\cite{de2012improving,xu2024large}. We aim to predict the water depth in the 2D SWE.

We use FloodCastBench~\cite{xu2025floodcastbench} to evaluate surrogate models, including FNO, PCNO, and DiffPCNO (see Fig.~\ref{fig:4}a for model configuration), for large-scale, cross-regional, and downscaled flood forecasting. 
Training data in FloodCastBench are generated by discretizing the SWE into a discrete domain using a finite difference (FD) method, a widely accepted hydrodynamic technique.
The dataset comprises four large-scale floods: Pakistan flood (18-31 August 2022, 85,616.5 $\mathrm{km^{2}}$),  Mozambique flood (16-20 March 2019, 6,190.9 $\mathrm{km^{2}}$), Australia flood (20 February-2 March 2022, 1,361.3 $\mathrm{km^{2}}$), and UK flood (4-7 December 2015, 135.5 $\mathrm{km^{2}}$). To assess the effectiveness and transferability of these models, we define two scenarios: low-fidelity forecasting using the Pakistan and Mozambique flood datasets (480 m spatial, 5 min temporal resolution) and high-fidelity forecasting using the Australia and UK flood datasets (60 m or 30 m spatial, 5 min temporal resolution). In each scenario, two long-term forecasting experiments are conducted using different training time steps. Further details are provided in the Supplementary Notes 3.

For long-term spatiotemporal flood forecasting, PCNO, which embeds momentum conservation in the output space via invariance, consistently outperforms FNO in both predictive accuracy and transferability. 
Ablation experiments (Fig.~\ref{fig:4}b) comparing FNO and PCNO with and without rainfall and terrain DEM demonstrate that incorporating these physical variables significantly enhances long-term flood predictability, yielding lower nRMSE values. 
The critical success index (CSI; defined in Methods) across flood depths (Extended Data Fig.~\ref{fig:e1}a) further confirms that embedding physical variables enhances forecasting, particularly for high water levels (exceeding 0.5 m). These results indicate that incorporating physical constraints in the input or output space effectively strengthens model predictability. Moreover, these surrogate models significantly outperform conventional hydrodynamic methods in computational efficiency. For instance, a two-day Pakistan flood simulation at 480 m resolution takes approximately one day using the hydrodynamic method on an NVIDIA A6000 GPU, whereas PCNO completes the same forecast in under one minute (Supplementary Table 6).

 \begin{figure}[htp!]
	\centering
	{\includegraphics[width = 0.9\textwidth]{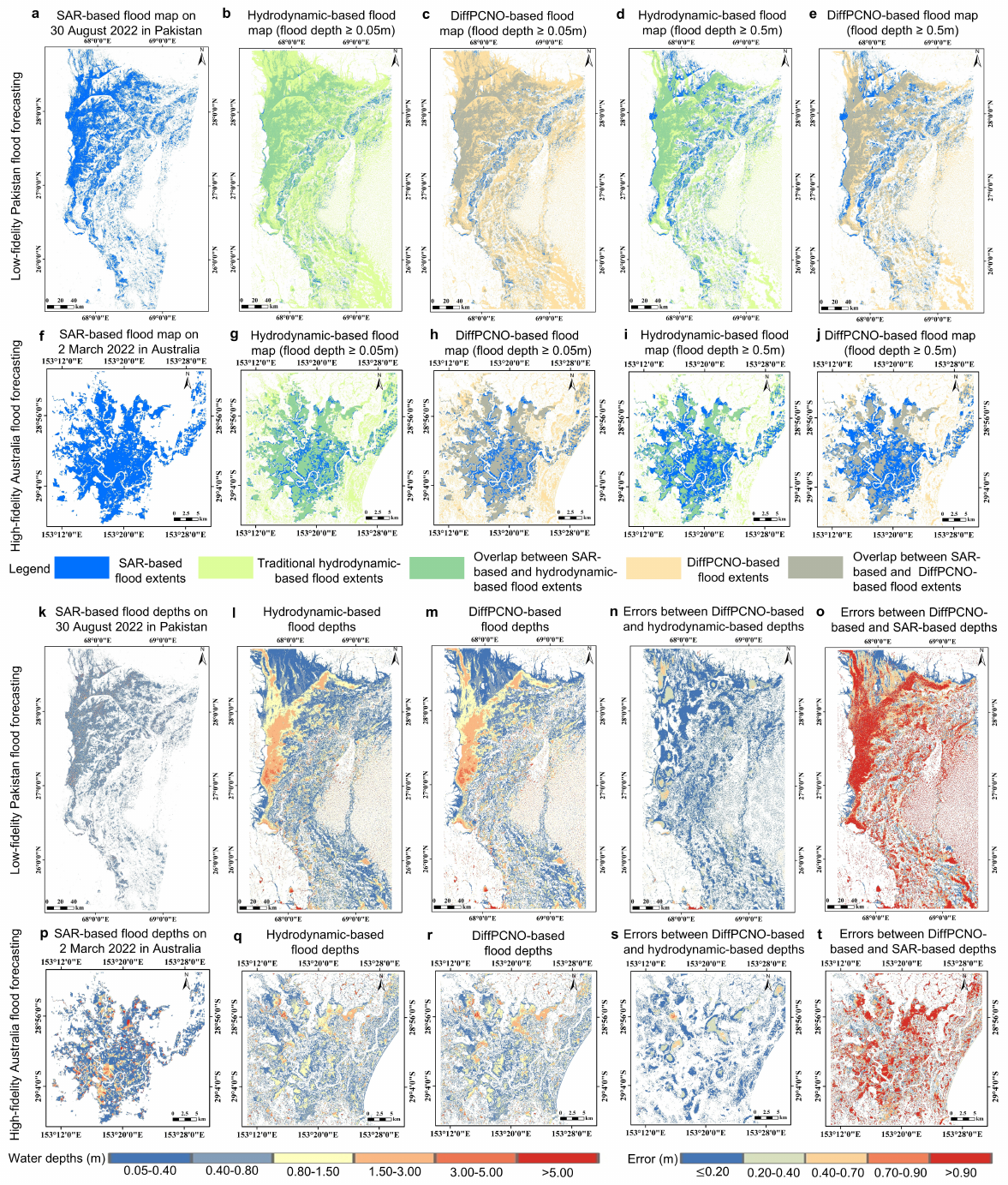}}
	\caption{Spatial variability of final flood inundation extents and depths across different methods. \textbf{a} and \textbf{f,} SAR-based flood maps for Pakistan on 30 August 2022 and Australia on 2 March 2022.  \textbf{b} and \textbf{g,} Traditional hydrodynamic-based flood maps (flood depth$\ge$0.05 m). \textbf{c} and \textbf{h,} DiffPCNO-based flood maps (flood depth$\ge$0.05 m). \textbf{d} and \textbf{i,} Traditional hydrodynamic-based flood maps (flood depth$\ge$0.5 m). \textbf{e} and \textbf{j,} DiffPCNO-based flood maps (flood depth$\ge$0.5 m). Compared with SAR-based maps, hydrodynamic- and DiffPCNO-based maps slightly overestimate flood extent at low water levels ($\ge$0.05 m), likely because water depths do not surpass certain land-cover heights (e.g., crops), yielding minimal SAR backscatter changes. \textbf{k} and \textbf{p,} SAR-based flood depths for Pakistan on 30 August 2022 and Australia on 2 March 2022. SAR-based flood depths are extracted using the DEM and SAR-based flood extents~\cite{cohen2018estimating}. \textbf{l} and \textbf{q,} Traditional hydrodynamic-based flood depths (flood depth$\ge$0.05 m). \textbf{m} and \textbf{r,} DiffPCNO-based flood depths (flood depth$\ge$0.05 m).
    \textbf{n} and \textbf{s,} Spatial distribution of errors between DiffPCNO-based and hydrodynamic-based flood depths. \textbf{o} and \textbf{t,} Spatial distribution of errors between DiffPCNO-based and SAR-based flood depths.
	}
	\label{fig:5}
\end{figure}
 We evaluate model performance under low- and high-fidelity flood forecasting tasks (Fig.~\ref{fig:4}b and Fig.~\ref{fig:4}c). All models achieve skillful long-term forecasts (nRMSE $<$ 0.1). PCNO and DiffPCNO exhibit lower nRMSE than FNO, indicating superior long-term forecasting performance. In transferability tests (Fig.~\ref{fig:4}d and Fig.~\ref{fig:4}e), PCNO and DiffPCNO maintain robust performance, particularly for extended forecasts over 48 continuous hours. Furthermore, embedding uncertainty via DiffPCNO does not compromise the spatiotemporal forecasting performance and transferability of PCNO. We further assess the models’ performance under varying training and test time steps. Comparing identical training time steps across different test time steps reveals that longer test durations (i.e., 48 hours) result in reduced performance compared to shorter test durations (i.e., 24 hours). Figs.~\ref{fig:4}f and g further indicate that splitting DiffPCNO’s 48-hour forecast into two 24-hour sequential forecasts substantially reduces rollout MSE in the second sequence, demonstrating the benefit of incorporating intermediate observations during long-term forecasts. Additionally, at a fixed test horizon, models trained with 24-hour steps outperform those trained with 12-hour steps, as the 24-hour training interval better aligns with the test horizons. 

High-resolution numerical grids can accurately capture complex flooding behaviors; however, traditional hydrodynamic approaches incur substantial computational costs~\cite{fraehr2023supercharging}.  In contrast, PCNO and DiffPCNO demonstrate strong resolution invariance, enabling zero-shot downscaling from 60 m training resolution to 30 m testing resolution with lower nRMSE than FNO (Figs.~\ref{fig:4}c and e), and maintaining superior downscaling accuracy across varying water depths (Extended Data Figs.~\ref{fig:e1}c and d). Furthermore, DiffPCNO captures uncertainty by leveraging a diffusion model. 
Specifically, for both low- and high-fidelity forecasts across spatiotemporal and cross-scenario settings (Figs.~\ref{fig:4}f-i), rollout MSE increases with forecast time, accompanied by progressively widening uncertainty bands.

We further analyze the spatial variability of final flood inundation extents and depths (Fig.~\ref{fig:5} and Extended Data Fig.~\ref{fig:6}) across the flood measurement data (synthetic aperture radar (SAR) and surveyed flood outlines), traditional hydrodynamic (FD) methods, and DiffPCNO. Overall, the inundation boundaries (Figs.~\ref{fig:5}a-j) from traditional hydrodynamic and DiffPCNO demonstrate a substantial degree of spatial consistency across different flood depths. Additionally, both approaches encompass a significant portion of the SAR-based maps, aligning well with the observed flood extents. Comparison of hydrodynamic-based flood depths (Figs.~\ref{fig:5}l and q) with DiffPCNO-based predictions (Figs.~\ref{fig:5}m and r) shows strong agreement in their spatial distributions, with error distributions (Figs.~\ref{fig:5}n and s) largely below 0.20 meters. In contrast, errors between DiffPCNO-based predictions and SAR-based flood depths from the same dates frequently exceed 0.9 m, reflecting the high uncertainty inherent in SAR-based depths due to the inherent inaccuracies of the DEM and estimated inundation extents. In transferability experiments (Extended Data Fig.~\ref{fig:6}), flood maps generated by DiffPCNO exhibit close agreement with hydrodynamic models at low water levels ($\ge$0.05 m), whereas at higher levels ($\ge$0.5 m) DiffPCNO systematically predicts greater extents, aligning more accurately with SAR observations and surveyed flood outlines. DiffPCNO also consistently estimates higher flood depths, likely reflecting its training on source-domain data capturing continuous depth increases under persistent rainfall.

\subsection{Atmospheric modeling}
\begin{figure}[h!]
	\centering
	{\includegraphics[width = 1.01\textwidth]{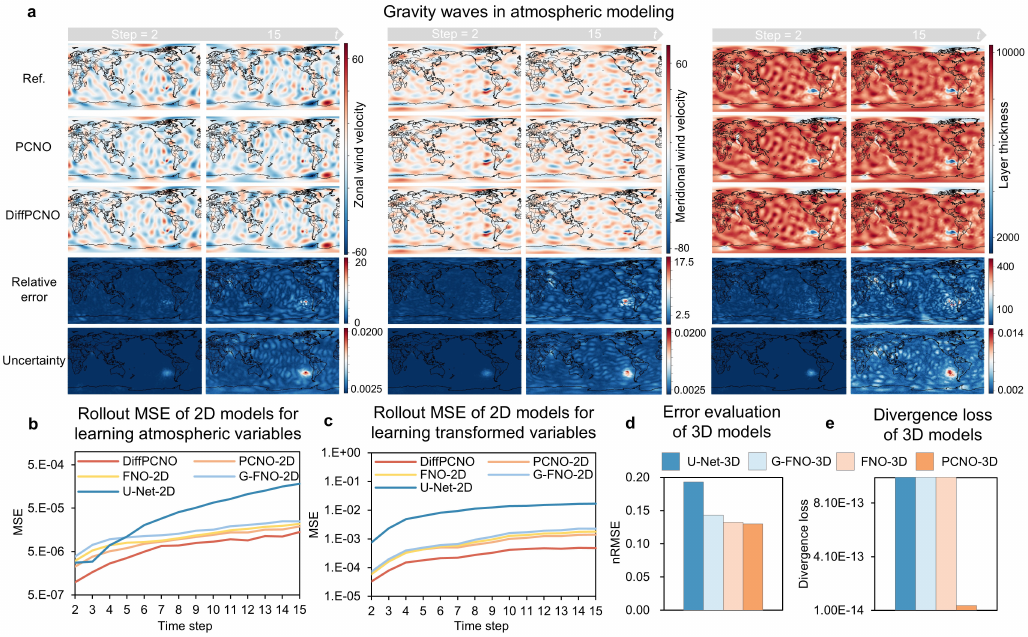}}
	\caption{Results of the atmospheric modeling. \textbf{a,} Visualization and uncertainty quantification using 2D models for learning atmospheric variables $(u_x, u_y, h)$ (15 min temporal, T63 spectral truncation, 15 time steps). \textbf{b,}  Rollout MSE  of 2D models, including PCNO-2D, DiffPCNO, and baseline methods for learning atmospheric variables. The $y$-axis is plotted on a logarithmic scale of the MSE. The 2D model employs a Markov training strategy, using spatial coordinates and initial conditions as input parameters. \textbf{c,}  Rollout MSE  of 2D models for learning  the transformed variables $(u_x h, u_y h \sin \theta, R h \sin \theta)$. The $y$-axis is plotted on a logarithmic scale of the MSE. \textbf{d,} Relative error of 3D models, including PCNO-3D and baseline methods. The 3D model employs a one-shot training strategy that incorporates spatial coordinates, temporal domain, and initial conditions. \textbf{e,} Divergence loss used to quantify mass conservation in 3D models. All models generate rollouts of $T = 14$ time steps (3.5 hours) conditioned on the first step ($T_{in} = 1$). 
	}
	\label{fig:7}
\end{figure}
General circulation models (GCM) numerically integrating the NS equations for the atmosphere and oceans form the foundation of modern weather and climate prediction. Recent advances in machine learning have introduced promising alternatives for atmospheric modeling~\cite{zhang2023skilful,kochkov2024neural,price2025probabilistic}. In the approximation for negligible depth compared to horizontal scales, the shallow water equations for simulating gravity waves in the Earth’s atmosphere are derived from the conservation of mass and momentum (see Eq. 2 in Supplementary Notes 4).
The GCM generates hourly outputs of gravity waves using a time step of $\Delta t = 15 \mathrm{~minutes}$ at T63 spectral truncation ($1.875^{\circ}$ horizontal resolution). 
We develop two-dimensional (2D) surrogate models to directly predict the layer thickness $h$, zonal wind velocity $u_x$, and meridional wind velocity $u_y$. To ensure divergence-free predictions, the 2D and three-dimensional (3D) surrogate models are trained to learn the transformed variables $(u_x h, u_y h \sin \theta, R h \sin \theta)$, with $R$ denoting the radius and $\theta$ the latitude. 
Further details are provided in the Supplementary Notes 4.

For 2D models, as the output lacks a temporal dimension (precluding gradient computation over time), PCNO-2D enforces momentum conservation in the output space through invariance. This incorporation of physical invariance (PCNO-2D) significantly enhances spatiotemporal forecasting performance for both atmospheric variables and their transformed forms, compared with baselines, including FNO-2D, G-FNO-2D, and U-Net-2D (Fig.~\ref{fig:7}b and Fig.~\ref{fig:7}c). Moreover, embedding uncertainty within the DiffPCNO framework further improves the spatiotemporal forecasting capability of the PCNO model (Fig.~\ref{fig:7}b and Fig.~\ref{fig:7}c). PCNO-3D enforces both momentum conservation through invariance and mass conservation through divergence-free conditions, achieving the lowest nRMSE among 3D baselines (U-Net-3D, G-FNO-3D, FNO-3D) (Fig.~\ref{fig:7}d).  The divergence loss (Fig.~\ref{fig:7}e) demonstrates that PCNO-3D effectively preserves mass in weather forecasting, with divergence values close to zero compared with other models.

DiffPCNO explicitly models the uncertainty of spatiotemporal forecasts via the consistency model. As shown in Fig.~\ref{fig:7}a and Supplementary Fig. 2, the uncertainty of atmospheric variables increases over time. Furthermore, the spatial patterns of uncertainty obtained by DiffPCNO align closely with the spatial distribution of relative forecast errors, providing a reliable measure for the confidence of spatiotemporal predictions in atmospheric modeling.

\section{Discussion}\label{sec12}
We have introduced the PCNO, which rigorously enforces physical constraints by projecting surrogate model outputs onto function spaces that satisfy predefined conservation laws. A physics-consistent projection layer efficiently implements momentum and mass conservation through transformations in Fourier space and can be seamlessly integrated into diverse surrogate models.  Building upon deterministic predictions, we further developed a probabilistic, DiffPCNO that leverages a consistency model to quantify and mitigate predictive uncertainties, thereby improving both the accuracy and reliability of long-term forecasts.
Across a range of benchmark systems, from turbulent flows to flood prediction and modeling of atmospheric waves, PCNO and DiffPCNO consistently deliver high-accuracy spatiotemporal predictions while preserving physical consistency and quantifying uncertainty. 
Notably, experiments on the flood forecasting dataset demonstrate that PCNO and DiffPCNO maintain high precision in cross-regional and large-scale flood forecasting.  
PCNO and DiffPCNO show potential for rapid flood forecasting under climate change through their strong generalization. 

However, the applicability of PCNO's mass and momentum conservations needs thorough verification across a broader range of PDEs. Additionally, PCNO has not explored other physical laws, such as energy conservation, nor has it been extensively tested with non-fluid spatiotemporal dynamics, which will be the focus of future research and development. Beyond the proposed DiffPCNO and PCNO-Refiner methods, further exploration of integrating probabilistic and deterministic learning is needed to develop a unified uncertainty quantification methodology for spatiotemporal dynamics via diffusion models, particularly leveraging the fast-generated consistency model. Moreover, directly embedding physical laws into diffusion models to ensure physically consistent generation remains a significant challenge for DiffPCNO.




\section{Methods}\label{sec3}
\subsection{Overview}
In the following, we provide a detailed explanation of the two-stage learning paradigm:  the deterministic learning through PCNO for embedding physics to learn spatiotemporal dynamics, and the diffusion model-based probabilistic learning for embedding uncertainty to learn spatiotemporal dynamics. Specifically, the proposed PCNO is developed by enforcing stringent physical consistency constraints, including the conservation of momentum and mass, within the framework of operator learning. Then, leveraging the rapid generation of the consistency model, a generative residual correction mechanism (DiffPCNO) is integrated to effectively quantify the inherent uncertainties present in PCNO predictions. All symbols are defined in Supplementary Table 1.
\subsection{Embedding physics to learn spatiotemporal dynamics processes }
Given a spatiotemporal dynamical system described by a set of nonlinear, coupled PDEs as,
\begin{equation}
\mathbf{u}_t(\mathbf{x}, \mathbf{t})=F\left(\mathbf{x}, \mathbf{t}, \mathbf{u}, \nabla_{\mathbf{x}} \mathbf{u}, \mathbf{u} \cdot \nabla_{\mathbf{x}} \mathbf{u}, \nabla^2 \mathbf{u}, \cdots\right),
\label{eq1}
\end{equation}
where $\mathbf{u}(\mathbf{x}, \mathbf{t}) \in \mathbb{R}^m$ represents the state variable with $m$ components defined over the spatiotemporal domain $\{(\mathbf{x}, \mathbf{t})\} \in \Omega \times \mathcal{T}$ . Here, $\Omega \subset \mathbb{R}^d$ and $\mathcal{T} \subset \mathbb{R}$ denote the $d$-dimensional spatial and temporal domain, respectively. $\nabla_{\mathbf{x}}$ is the Nabla operator with respect to the spatial coordinate $\mathbf{x}$, and $F(\cdot)$ is a nonlinear function describing the right-hand side of PDEs. The solution to this problem is governed by the initial condition  $\mathcal{I}(\mathbf{u}; t=0, \mathbf{x} \in \Omega)=0$, and the boundary condition $\mathcal{B}\left(\mathbf{u}, \nabla_{\mathbf{x}} \mathbf{u}, \cdots ; \mathbf{x} \in \partial \Omega\right)=0$, where $\partial \Omega$ represents the boundary of the system domain. These spatiotemporal dynamical systems 
are often at least partially constrained by well-known fundamental laws, such as conservation laws and symmetries. A pure data-driven paradigm is to approximate the PDE from its solution samples, which neglects these intrinsic fundamental physical laws in the data. As a result, the performance of such models is heavily contingent upon the quantity and diversity of the available data.  To enhance the efficacy and robustness of spatiotemporal predictions, our objective is to incorporate these fundamental physical laws directly at the architectural level of surrogate modeling, thereby ensuring not only physically consistent predictions despite limited training data but also reliable long-term forecasts of the system's dynamic behavior for any given initial conditions.

Borrowing the concepts of numerical discretization~\cite{long2018pde,rao2023encoding}, the design principle of the proposed PCNO is rooted in the observation that the state variable $\mathbf{u}$ can be updated iteratively from one time step $\mathbf{u}_{t}$ to the next $\mathbf{u}_{t+1}$. That said, the state variable $\mathbf{u}$ would be updated by,
\begin{equation}
\mathbf{u}_{t+1}=\mathcal{G}\left[\mathbf{u}_{t} ; \theta\right](\mathbf{x}),
\label{eq2}
\end{equation}
where $\mathbf{u}_{t+1}$ is the predicted variable/solution at time $t+1$, and $\mathcal{G}$ denotes  any NO surrogate parameterized by $\theta$ that combines a series of operations to compute $F(\cdot)$ in Eq.~\ref{eq1}. The structure of Eq.~\ref{eq2} facilitates the design of an NO for long-term spatiotemporal predictions. 
\subsubsection{Surrogate model}
Although our surrogate model of spatiotemporal dynamics
processes is independent of any specific NO architecture, for illustration, we consider the generic integral neural operators~\cite{li2020fourier,azizzadenesheli2024neural} as the surrogate model. The  NO $\mathcal{G}$ is  to learn the underlying mapping from $\mathbf{u}_{t}$ to $\mathbf{u}_{t+1}$, denoted as,
\begin{equation}
\mathcal{G}\left[\mathbf{u}_{t} ; \theta\right](\mathbf{x}):=\mathcal{Q} \circ \mathcal{J}_L \circ \cdots \circ \mathcal{J}_1 \circ \mathcal{P}[\mathbf{u}_{t}](\mathbf{x}),
\label{eq4}
\end{equation}
where the surrogate model $\mathcal{G}$ consists of sequential steps that first lift the input channel using $\mathcal{P}$, then apply $L$-layer nonlinear operators $\left\{\mathcal{J}_1, \mathcal{J}_2, \ldots, \mathcal{J}_L\right\}$, and finally project back to the output function space using $\mathcal{Q}$. Both  $\mathcal{P}$ and $\mathcal{Q}$ are pixel-wise transformations that can be implemented using a multilayer perceptron (MLP). 
The nonlinear operator $\mathcal{J}_l$ consists of a local linear transformation operator $W_l$, an integral kernel operator $\mathcal{K}_l$, and an activation function $\sigma$. The architectures of NOs primarily differ in the design of the update rules for their nonlinear operator layers. In the context of spatiotemporal problems defined on a structured domain 
$\Omega$, the Fourier Neural Operator (FNO)~\cite{li2020fourier} is commonly employed, where the integral kernel operators $\mathcal{K}_l$ are linear transformations in the frequency domain (Fourier integral kernel operator).
Specifically, for an input feature function $v_l$ at the $l$th layer, $\mathcal{J}_l: v_l \rightarrow v_{l+1}$ is computed as follows,
\begin{equation}
\begin{aligned}
v_{l+1}&=\sigma(W_l v_l(\mathbf{x})+\left(\mathcal{K}\left(v_l\right)\right)(\mathbf{x})), \\
\text{Fourier integral kernel operator:}&\quad  \left(\mathcal{K}\left(v_l\right)\right)(\mathbf{x})=\mathcal{F}^{-1}\left(R_\phi \cdot \mathcal{F}\left(v_l\right)\right)(\mathbf{x}),
\end{aligned}
\end{equation}
where the Fourier integral kernel operator $\mathcal{K}$ is processed sequentially through three operations: the Fast Fourier Transform (FFT) $\mathcal{F}$, the Fourier transform of a periodic function $R_\phi$ parameterized by $\phi$, and the inverse FFT $\mathcal{F}^{-1}$. $R_\phi$  is directly parameterized as a complex-valued tensor that maps to the values of the appropriate Fourier modes $k$, and the values of $R_\phi$ are learned from training data.  

FNO offers several advantages, including resolution invariance (zero-shot downscaling) through Fourier integral kernel operators, flexible embedding of physical constraints, and improved generalizability across diverse input scenarios. However, the output of the surrogate model 
$\mathcal{G}$ (such as FNO) may not fully adhere to the physical laws within the domain. To address this, we propose a mechanism to modify $\mathcal{G}$ and derive a new physics-consistent surrogate model through a physics-consistent projection layer $\mathcal{D}$. 

\subsubsection{Physics-consistent projection layer}
Given a set of physical constraints $\mathcal{C}$ (such as mass and momentum conservations), the physics-consistent projection layer $\mathcal{D}$ is defined as a general framework for projecting the output of any operator surrogate model onto the space of functions that satisfy these physical constraints. That is,
\begin{equation}
\mathbf{u}_{t+1}=\mathcal{D} \circ \mathcal{G}\left[\mathbf{u}_t ; \theta\right](\mathbf{x}), \quad  \mathcal{D} \text { subject to } \mathcal{C},
\end{equation}
where 
$\mathcal{D}$ ensures that the output adheres to the prescribed physical laws. The construction of $\mathcal{D}$ depends on how the physical constraints $\mathcal{C}$ are efficiently projected in the operator surrogate model. Instead of directly enforcing these constraints in physical space, where the computation of derivatives can be computationally intensive, we employ a Fourier transform  $\mathcal{F}$ to impose the constraints in Fourier space, subsequently transforming back to the physical space (Fig.~\ref{fig:1}b). Fourier space enhances computational efficiency by enabling gradient calculations in physical constraints through multiplication operations, thus bypassing finite difference approximations and ensuring a more stable and efficient numerical solution~\cite{duruisseaux2024towards,liu2024harnessing}. Furthermore, when the domain discretization is uniform, $\mathcal{F}$ can be replaced with the FFT to enhance computational efficiency further. Thus, the physics-consistent projection layer $\mathcal{D}$ in Fourier space is defined as,
\begin{equation}
\mathbf{u}_{t+1}=\mathcal{F}^{-1} (\mathcal{D}^* \circ \mathcal{F}(\mathcal{G}\left[\mathbf{u}_t ; \theta\right]))(\mathbf{x}),
\end{equation}
where $\mathcal{D}^*$ represents the imposition of physical constraints in Fourier space. Specifically, by applying $\mathcal{D}^*$, $\mathcal{F}(\mathcal{G}\left[\mathbf{u}_t; \theta\right])$ is transformed into a new Fourier space of functions that satisfy these physical constraints.  In this work, we consider the conservation of mass and momentum in spatiotemporal dynamics as a representative application of the physics-consistent projection layer. These two conservation laws are fundamental in fields such as flood and climate modeling.

\textbf{Expressing momentum conservation in the projection layer.}
Noether's theorem establishes a fundamental link between symmetries and conservation laws in physical systems~\cite{noether1971invariant, bluman2010applications}. Specifically, translational invariance of the physical model ensures the conservation of linear momentum, while invariance under rotations guarantees the conservation of angular momentum. A detailed introduction to Noether's theorem, along with a comprehensive proof of the symmetries that lead to these conservation laws, is provided in the Supplementary Notes 5.
By leveraging the relationships between symmetries and conservation laws, we propose a data-driven projection layer that enforces momentum conservation by embedding translational and rotational invariances into the physical constraints. 

\textbf{Momentum-conserving projection.} To derive the momentum-conserving projection layer with invariance properties, we exploit the connection between symmetry embedding in both the frequency and physical domains. Specifically, applying a transformation to a function in physical space induces the same transformation on its Fourier transform~\cite{helwig2023group}.
For translation invariance, the Fourier layer serves as a translation-invariant projection, inherently preserving linear momentum without the need to impose physical constraints explicitly (see Supplementary Notes 6.1 for a detailed proof). Rotation invariance in the output space is achieved by configuring the projection operator to implement a rotation-invariant convolution in the frequency domain through a designed rotation-invariant kernel.
 Such a kernel can be constructed by parameterizing only half of the complex-valued weights along a designated axis and generating the remaining half through a symmetric rotation about the kernel's center. This design enforces rotational invariance in the learned representations (see Supplementary Notes 6.2 for a detailed proof).

To further enhance the expressiveness and invariance of the momentum-conserving projection layer, invariant convolutions $W_{inv}$ with rotational symmetry~\cite{cohen2016group} are applied after the projection, without introducing additional learnable parameters. Moreover, a residual connection using invariant convolutions is introduced to effectively retain information from the output space of the surrogate model. The momentum-conserving projection $\mathcal{D}_{mom}$ (Fig.~\ref{fig:1}c) is defined as,
\begin{equation}
\mathcal{D}_{mom}=W_{inv} \mathcal{G}\left[\mathbf{u}_t; \theta\right]+W_{inv}\left(\mathcal{F}^{-1}\left(\left(L_R[\mathcal{F}(W)]\right) \cdot \mathcal{F}(\mathcal{G}\left[\mathbf{u}_t; \theta\right])) \right)\right),
\label{Eq.9}
\end{equation} 
where $L_R[\mathcal{F}(W)]$ is the rotation-invariant kernel we aim to learn.
The implementation details of the momentum-conserving projection, along with a rigorous proof of its invariance and momentum-preserving properties, are provided in the Supplementary Notes 6.

\textbf{Expressing mass conservation in the projection layer.} Mathematically, mass conservation can be represented by a continuity equation, which characterizes the relationship between the quantity of a substance and its corresponding transport, 
\begin{equation}
\frac{\partial \rho}{\partial \mathbf{t}}(\mathbf{x}, \mathbf{t})+\nabla_{\mathbf{x}} \cdot \mu(\mathbf{x}, \mathbf{t})=0,
\label{eq11}
\end{equation}
where $\rho$  represents the fluid density, and $\mu$ denotes the mass flux. $\mu=\rho \mathbf{u}$, where  $\mathbf{u}$ is the velocity field. Mass conservation (Eq.~\ref{eq11})  can be reformulated as a divergence-free vector field by incorporating both spatial and temporal dimensions, leading to the condition, $\operatorname{div}\binom{\rho}{\rho \mathbf{u}}=0$. Therefore, we propose modeling solutions of the mass conservation in the projection layer by the divergence-free condition. 

\textbf{Mass-conserving projection.} To construct the mass-conserving projection, we illustrate the procedure through a (quasi-)static case where the density  $\rho$ is time-independent. In this case,  the divergence operator is applied only to the spatial variables. The output field is defined as a 2D velocity field
$\mathbf{u}(\mathbf{x})=(\mathbf{u}_{1}(\mathbf{x}),\mathbf{u}_{2}(\mathbf{x}))$, where $\mathbf{x}=(\mathbf{x}_1, \mathbf{x}_2) \in \mathbb{R}^2$. The corresponding divergence-free condition is  expressed as,
\begin{equation}
\operatorname{div}(\mathbf{u})=\nabla \cdot \mathbf{u}\left(\mathbf{x}_1, \mathbf{x}_2\right)=\frac{\partial \mathbf{u}_1}{\partial \mathbf{x}_1}+\frac{\partial \mathbf{u}_2}{\partial \mathbf{x}_2}=0 .
\label{eq12}
\end{equation}
Assuming that $\mathbf{u}$ is a 2D periodic function, its non-zero Fourier modes $k = (k_1, k_2)$ belong to $\mathbb{Z}^2$. Taking the Fourier transform of Eq.~\ref{eq12} results in,
\begin{equation}
\mathcal{F}(\frac{\partial \mathbf{u}_1}{\partial \mathbf{x}_1}+\frac{\partial \mathbf{u}_2}{\partial \mathbf{x}_2})=0 \quad \Longrightarrow \quad k_1 \tilde{\mathbf{u}}_1(k)+k_2 \tilde{\mathbf{u}}_2(k)=0,
\end{equation}
where $\tilde{\mathbf{u}}$ denotes the Fourier transform of 
$\mathbf{u}$. This divergence-free condition in Fourier space must be satisfied for all $\left(k_1, k_2\right) \in \mathbb{Z}^2$. Hence, the key to constructing a data-driven projection layer lies in ensuring that $\tilde{\mathbf{u}}$ is divergence-free in Fourier space.
Inspired by the postprocessing to eliminate divergences~\cite{cao2025spectralrefiner,liu2024harnessing}, we employ a discrete Helmholtz decomposition~\cite{girault2012finite} in the frequency domain to project the output of surrogate models onto a divergence-free field $\tilde{\mathbf{u}}$,
\begin{equation}
\mathcal{C}_{div}^*(\mathcal{F}(\mathcal{G})) = \tilde{\mathbf{u}}=\mathcal{F}(\mathcal{G}) -\frac{\nabla(\nabla \cdot \mathcal{F}(\mathcal{G}))}{\Delta},
\end{equation}
where $\mathcal{C}_{div}^*$ is the divergence-free condition in Fourier space. $\nabla(\nabla \cdot \mathcal{F}(\mathcal{G}))$ represents the gradient of the divergence of $\mathcal{F}(\mathcal{G})$. $\Delta$  denotes the Laplacian in the frequency domain. A detailed derivation of this project is provided in the Supplementary Notes 7.1. Furthermore, the mass-conserving projection is also applicable to time-dependent systems, such as atmospheric modeling based on the SWE, with a detailed proof presented in the Supplementary Notes 7.2.

To improve the representability and learnability of the mass-conserving projection, a spectral convolution $W_{spe}$ is used following the Fourier transform of the output $\mathcal{F}(\mathcal{G})$. Thus, the mass-conserving projection $\mathcal{D}_{mass}$ (Fig.~\ref{fig:1}d) is represented as,
\begin{equation}
\mathcal{D}_{mass}=\mathcal{F}^{-1} (\mathcal{C}_{div}^*\left( W_{spe}(\mathcal{F}(\mathcal{G}\left[\mathbf{u}_t ; \theta\right]))\right).
\end{equation}

\subsubsection{Physics-consistent neural operator}
With the above analysis, we develop a PCNO architecture that ensures physics-consistent outputs by modifying the surrogate model in Eq.~\ref{eq4} as follows,
\begin{equation}
\begin{aligned}
& \mathbf{u}_{t+1}= \mathcal{D} \circ \mathcal{Q} \circ \mathcal{J}_L \circ \cdots \circ \mathcal{J}_1 \circ \mathcal{P}[\mathbf{u}_{t}](\mathbf{x}), \\
 \mathcal{D}  & =  
\begin{cases}
\mathcal{D}_{\text{mom}}, & \text{if momentum conservation is enforced}; \\
\mathcal{D}_{\text{mass}}, & \text{if mass conservation is enforced}; \\
\mathcal{D}_{\text{mom}} \circ \mathcal{D}_{\text{mass}}, & \text{if both momentum and mass conservations are enforced}.
\end{cases}
\end{aligned}
\end{equation}
The proposed architecture is compatible with any surrogate model and can be seamlessly integrated into any neural network. In this work, we primarily employ the FNO as a representative surrogate model.
The physics-consistent projection layer acts as a flexible module, enabling the framework to accommodate a wide range of spatiotemporal dynamical systems (see details in Supplementary Notes 8.1). 
PCNO  utilizes a Markov training strategy for 1D and 2D spatiotemporal prediction, and a one-shot training strategy for 3D prediction tasks (see details in Supplementary Notes 8.2). 
The relative mean square error is utilized as the loss function for training PCNO.

\subsection{Embedding uncertainty to learn spatiotemporal dynamics processes}
To effectively quantify the uncertainties inherent in spatiotemporal dynamical processes, we integrate a generative residual correction mechanism based on a diffusion-based consistency model to estimate and mitigate predictive uncertainties, thereby enhancing both the accuracy and reliability of spatiotemporal forecasts, particularly for long-term predictions. Specifically, we formulate the probabilistic one-step-ahead forecast as,
\begin{equation}
\begin{aligned}
\mathbf{u}_{t+1}&=\mathcal{D} \circ \mathcal{G}\left[\mathbf{u}_t ; \theta\right]+\mathbf{r}_{t+1}, \\
\mathbf{r}_{t+1} & \sim \mathcal{P}\left[\mathbf{r} \mid \mathbf{u}_t, \mathcal{D} \circ \mathcal{G}\left[\mathbf{u}_t; \theta\right];\theta_g\right],
\end{aligned}
\end{equation}
where $\mathcal{P}$ denotes the generative consistency model parameterized by $\theta_g$, which refines the forecasting of PCNO and provides uncertainty quantification. The consistency model captures the probabilistic nature of the residuals from the deterministic model, representing a distribution conditioned on the current state $\mathbf{u}_t$ and deterministic prediction $\mathcal{D} \circ \mathcal{G}\left[\mathbf{u}_t; \theta\right]$. This formulation explicitly models the uncertainty and stochasticity inherent in physical processes that are not captured by the deterministic PCNO model.
By sampling the residual error $\mathbf{r}_{t+1}$, the model learns spatiotemporal trajectories from data. Beyond providing predictive uncertainties arising from intrinsic noise, these spatiotemporal trajectories enable the study of long-term behavior, stability, and scaling properties in spatiotemporal dynamical systems.
\subsubsection{Consistency models}
Generative models, particularly diffusion models, have achieved notable success across domains~\cite{ho2020denoising,song2021scorebased,price2025probabilistic,yang2025generative}, and have recently been adapted for spatiotemporal dynamics~\cite{li2024learning,du2024conditional}. Consistency models~\cite{song2023consistency} are a class of diffusion-based generative models that produce high-quality samples in a single step while retaining the flexibility of multi-step sampling to trade computational cost for fidelity. 
Unlike generative adversarial networks (GANs)~\cite{goodfellow2014generative}, they avoid adversarial training, and unlike score-based diffusion models~\cite{ho2020denoising,song2021denoising}, they generate samples without iterative denoising.

Consistency models can be trained using either consistency distillation (CD) or consistency training (CT). CD involves pre-training a diffusion model and distilling its knowledge into a consistency model, but at the cost of increased computational overhead. CT, on the other hand, trains the model directly from data, forming a standalone generative model. Here, we focus on CT to enhance efficiency and accuracy in spatiotemporal forecasting.

The foundation of consistency models is the probability flow ordinary differential equation (ODE)~\cite{song2021scorebased}, which defines a bijective mapping between noisy and clean data samples. The model learns this mapping through a consistency function $\boldsymbol{f}(\boldsymbol{x}, t)$, that is,
\begin{equation}
\boldsymbol{f}(\boldsymbol{x}_t, t)=\boldsymbol{f}\left(\boldsymbol{x}_t^{\prime}, t^{\prime}\right), \quad \forall t, t^{\prime} \in\left[t_{min}, t_{max}\right],
\end{equation}
where the time interval is defined as $t_{min}=0.002$ and $t_{max}=80$~\cite{song2023consistency}, and the consistency function is constrained by the boundary conditions $\boldsymbol{f}\left(\boldsymbol{x}_\epsilon, \epsilon\right)=\boldsymbol{x}_\epsilon$. The consistency model, $\boldsymbol{f}(\boldsymbol{x}, t;\boldsymbol{\theta})$, is implemented as a neural network trained to approximate the target consistency function $\boldsymbol{f}(\boldsymbol{x}, t)$. Following~\cite{song2023consistency}, we parameterize the model using skip connections,
\begin{equation}
\boldsymbol{f}(\boldsymbol{x}, t;\boldsymbol{\theta})=c_{\text {skip }}(t) \boldsymbol{x}+c_{\text {out }}(t) F(\boldsymbol{x}, t;\boldsymbol{\theta}),
\end{equation}
where $\boldsymbol{F}(\boldsymbol{x}, t;\boldsymbol{\theta})$ is a U-Net, while $c_{\text {skip }}(t)$ and $c_{\text {out }}(t)$ are differentiable functions such that $c_{\text {skip }}\left(t_{\min }\right)=1$ and $c_{\text {out }}\left(t_{\min }\right)=0$.

To train the consistency model, we discretize the probability flow ODE using a sequence of time steps $t_{\min }=t_1<t_2<\cdots<t_N=t_{\max}$. The discrete time step is determined as,
\begin{align}
t_i &= \left(t_{\min}^{1/\rho} + \frac{i-1}{N-1} \left(t_{\max}^{1/\rho} - t_{\min}^{1/\rho}\right) \right)^\rho, \\
i \sim p(i), & \quad
p(i) \propto 
\operatorname{erf}\left(\frac{\log(\sigma_{i+1}) - P_{\text{mean}}}{\sqrt{2} P_{\text{std}}}\right)
-
\operatorname{erf}\left(\frac{\log(\sigma_i) - P_{\text{mean}}}{\sqrt{2} P_{\text{std}}}\right)
\end{align}

where $\rho=7$, $P_{\text {mean }}=-1.1$, $P_{\text {std }}=2.0$, and $\operatorname{erf}$ denotes the error function. A random index $i$ is sampled from a discretized lognormal distribution $p(i)$ to enhance the sample quality of the consistency model. Furthermore,
CT is further enhanced using an improved discretization curriculum~\cite{song2024improved},
\begin{equation}
N(\mathrm{k})=\min \left(s_0 2^{\left\lfloor\frac{\mathrm{k}}{\mathrm{K}^{\prime}}\right\rfloor}, s_1\right)+1, \quad \mathrm{K}^{\prime}=\left\lfloor\frac{\mathrm{K}}{\log _2\left\lfloor s_1 / s_0\right\rfloor+1}\right\rfloor,
\end{equation}
where $\mathrm{k}$ denotes the current training step and $\mathrm{K}$ the total number of steps. The initial and maximum discretization steps are set to $s_0=10$ and $s_1=1280$, respectively. 

The model is trained by minimizing the consistency training
loss $\mathcal{L}_{\mathrm{CT}}$,
\begin{equation}
\mathcal{L}_{\mathrm{CT}}\left(\boldsymbol{\theta}, \boldsymbol{\theta}^{-}\right)=\mathbb{E}\left[\lambda\left(t_i\right) d\left(\boldsymbol{f}\left(\boldsymbol{x}+t_{i+1} \mathbf{z}, t_{i+1};\boldsymbol{\theta}\right), \boldsymbol{f}\left(\boldsymbol{x}+t_i \mathbf{z}, t_i;\boldsymbol{\theta}^{-}\right)\right)\right],
\label{eq23}
\end{equation}
where parameters $\boldsymbol{\theta}$ and $\boldsymbol{\theta}^{-}$ denote the student and teacher network weights, set to $\boldsymbol{\theta} = \boldsymbol{\theta}^{-}$~\cite{song2024improved}. The weighting function is $\lambda\left(t_i\right)=\frac{1}{t_{i+1}-t_i}$. $d(x, y)$ is defined using the Pseudo-Huber metric, which smoothly interpolates between $\ell_1$ and squared $\ell_2$ norms,
\begin{equation}
d(x, y)=\sqrt{\|x-y\|_2^2+c^2}-c,
\end{equation}
where $c>0$ is an adjustable constant.

Thus, the improved consistency model~\cite{song2024improved} is employed for DiffPCNO and trained in a self-supervised manner. After the CT, samples are generated by initializing with noise $\mathbf{z}$ and computing $\boldsymbol{x}=\boldsymbol{f}\left(\mathbf{z}, t_{max};\boldsymbol{\theta}\right)$; the model further supports multistep generation.
\subsubsection{Diffusion model-enhanced PCNO}
DiffPCNO corrects its prediction residuals through the consistency model by conditioning on the deterministic prediction $\mathbf{\hat{u}}_{t+1}=\mathcal{D} \circ \mathcal{G}\left[\mathbf{u}_t; \theta\right]$ from PCNO and the current state $\mathbf{u}_t$. Its objective is to capture the residual distribution $\mathbf{y}-\mathbf{\hat{u}}_{t+1}$, where 
$\mathbf{y}$ denotes the ground-truth solution at $t+1$. Combined with the consistency training loss in Eq.~\ref{eq23}, the training loss of DiffPCNO is formulated as,
\begin{equation}
\mathbb{E}\left[\lambda\left(t_i\right) d\left(\boldsymbol{f}\left((\mathbf{y}-\mathbf{\hat{u}}_{t+1})+t_{i+1} \mathbf{z}, t_{i+1}, \mathbf{\hat{u}}_{t+1}, \mathbf{u}_t;\boldsymbol{\theta}\right), \boldsymbol{f}\left((\mathbf{y}-\mathbf{\hat{u}}_{t+1})+t_i \mathbf{z}, t_i, \mathbf{\hat{u}}_{t+1}, \mathbf{u}_t;\boldsymbol{\theta}^{-}\right)\right)\right].
\label{eq25}
\end{equation}
We employ a U-Net architecture (see Supplementary Notes 9.3) to implement the CT within DiffPCNO. Detailed training and sampling procedures are provided in the Supplementary Notes 9.1.

To capture the inherent uncertainty in the prediction results, we further incorporate a consistency model-based refinement process (PCNO-Refiner) that enhances the initial outputs of PCNO. PCNO-Refiner conditions on the deterministic prediction $\mathbf{\hat{u}}_{t+1}$ from PCNO and the current state $\mathbf{u}_t$,  aiming to produce predictions closer to the ground truth $\mathbf{y}$. Its training loss is defined as,
\begin{equation}
\mathbb{E}\left[\lambda\left(t_i\right) d\left(\boldsymbol{f}\left(\mathbf{y}+t_{i+1} \mathbf{z}, t_{i+1}, \mathbf{\hat{u}}_{t+1}, \mathbf{u}_t;\boldsymbol{\theta}\right), \boldsymbol{f}\left(\mathbf{y}+t_i \mathbf{z}, t_i, \mathbf{\hat{u}}_{t+1}, \mathbf{u}_t;\boldsymbol{\theta}^{-}\right)\right)\right].
\label{eq26}
\end{equation}
Further details of PCNO-Refiner are provided in the Supplementary Notes 9.2. DiffPCNO and PCNO-Refiner employ a Markov training strategy for spatiotemporal process learning.

To evaluate the uncertainty quantification capability of our methods, we sample 50 trajectories for each test case. For each trajectory, a single prediction sample is generated at every autoregressive step and subsequently propagated to the next step, thereby enabling the iterative forecasting of the complete sequence.
\subsection{Evaluation metrics}
We use nRMSE~\cite{li2024learning,xu2024physics} as the primary evaluation metric, defined as,
\begin{equation}
\operatorname{nRMSE}(\mathbf{y}, \hat{\mathbf{y}})=\frac{1}{n} \sum_{i=1}^n \frac{\left\|\hat{\mathbf{y}}_i-\mathbf{y}_i\right\|_2}{\left\|\mathbf{y}_i\right\|_2},
\end{equation}
where $\hat{\mathbf{y}}_i$ and $\mathbf{y}_i$ are the predicted and true solutions of the $i$-th test sample. $n$ is the number of test samples and $\|\cdot\|_2$ is the $L_2$ norm. In addition, MSE~\cite{lippe2023pde}, defined as,
\begin{equation}
\mathrm{MSE}=\frac{1}{n} \sum_{i=1}^n\left\|\hat{\mathbf{y}}_i-\mathbf{y}_i\right\|_2^2,
\end{equation}
is used to quantify the magnitude of deviation between the predicted and true solutions.

For the KSE dynamic, we define a high-correlation time step to evaluate the stability and reliability of models in long-term forecasting. Specifically, we calculate the Pearson correlation coefficient $r$ between predicted and true values,
\begin{equation}
r = \frac{\sum_{i=1}^n (\mathbf{y}_i - \bar{\mathbf{y}})(\hat{\mathbf{y}}_i - \hat{\bar{\mathbf{y}}})}{\left\|\mathbf{y} - \bar{\mathbf{y}}\right\|_2 \left\|\hat{\mathbf{y}} - \hat{\bar{\mathbf{y}}}\right\|_2},
\end{equation}
where $\bar{\mathbf{y}}$ and $\hat{\bar{\mathbf{y}}}$
represent the mean values of the true and predicted solutions, respectively. The high-correlation time step is defined as the time at which the average correlation declines below 0.9 or 0.8.

To evaluate physical consistency, we introduce two complementary metrics: the divergence loss and the momentum loss. The divergence loss quantifies deviations from the divergence-free condition, thereby evaluating models’ ability to conserve mass,
\begin{equation} \mathcal{L}_{\mathrm{div}}=\frac{1}{N} \sum_{i=1}^{N} \left|\nabla \cdot \mathbf{u}_{\mathrm{pred}, i}\right|, 
\end{equation}
where  $N$ denotes the total number of spatial points, and $|\cdot|$ is the absolute value. $\nabla \cdot \mathbf{u}_{\mathrm{pred}, i}$ is the divergence of the predicted vector field at the $i$-th spatial point. The gradient is computed using FFT pseudo-spectral methods~\cite{xu2024large}.

The momentum loss~\cite{xu2024physics} measures the discrepancy between the total predicted and true momentum, thereby assessing models’ ability to conserve momentum,
\begin{equation}
\mathcal{L}_{\mathrm{M}} = \frac{1}{N} \Big\| \sum_{i=1}^{N} M_{\mathrm{pred}, i} - \sum_{i=1}^{N} M_{\mathrm{ref}, i} \Big\|_2^2,
\end{equation}
where where $M_{\mathrm{pred}, i}$ and $M_{\mathrm{ref}, i}$ represent the predicted and reference momentum at the $i$-th spatial point, respectively.

We also consider the critical success index (CSI) in flood forecasting~\cite{xu2025floodcastbench}, which measures the spatial accuracy of classifying cells as flooded or non-flooded for a given threshold $\gamma$, as follows,
\begin{equation}
\mathrm{CSI}=\frac{\mathrm{TP}}{\mathrm{TP}+\mathrm{FP}+\mathrm{FN}},
\end{equation}
where $\mathrm{TP}$ represents true positives (cells where both predictions and ground truths exceed $\gamma$ ), $\mathrm{FP}$ represents false positives (cells where ground truths are below $\gamma$ but predictions exceed $\gamma$ ), and $\mathrm{FN}$ represents false negatives (cells where the model fails to predict a flooded area). In our experiments, we set $\gamma \in\{0.05 \mathrm{~m}, 0.5 \mathrm{~m}\}$ to account for varying water depths.

\subsection{Baseline methods}
State-of-the-art neural numerical solvers are considered benchmarks. 
\textbf{U-Net:} A commonly employed architecture for image-to-image regression tasks, available in both 2D  and 3D  versions
~\cite{ronneberger2015u}.
\textbf{FNO:} 3D FNO with one-shot training strategy, utilizing direct convolutions in space-time~\cite{li2020fourier}. 2D FNO with a Markov training strategy in time~\cite{li2020fourier}. 
\textbf{G-FNO:} 3D group equivariant FNO with the group of translations and $90^{\circ}$ rotations (p4), utilizing one-shot training strategy in space-time~\cite{helwig2023group}. 2D group equivariant FNO with the group of translations and $90^{\circ}$ rotations (p4), utilizing a Markov training strategy in time~\cite{helwig2023group}. 
\textbf{PDE-Refiner:} Diffusion model-based multistep refinement process to learn spatiotemporal PDEs, utilizing a denoising diffusion probabilistic model with a Markov training strategy over time~\cite{lippe2023pde}. 
\textbf{PDE-Refiner+:} The deterministic prediction generated by PCNO is incorporated into the input of the PDE-Refiner~\cite{lippe2023pde}.
\textbf{Consistency model:} The improved consistency model~\cite{song2024improved}  employs consistency training with a Markov strategy over time, without incorporating the deterministic predictions generated by PCNO as inputs.
\textbf{ClawFNO:}  Conservation law–encoded FNOs inherently enforce mass conservation by generating divergence-free solution fields in the output space, where the divergence-free condition is rigorously imposed through an antisymmetry constraint~\cite{liu2024harnessing}. ClawFNO employs a Markov training strategy to propagate solutions over time.
\textbf{LNO:} Incorporating a Laplace layer in NOs to model specific dynamical systems~\cite{cao2024laplace}, with a Markov training strategy over time. 
\textbf{PCNO w/o Momentum:} The physics-consistent projection layer in PCNO is restricted to the enforcement of mass-conserving projection.
\textbf{PCNO w/o Mass:} The physics-consistent projection layer in PCNO focuses exclusively on the momentum-conserving projection.
\textbf{FNO without rainfall and terrain DEM:} In flood forecasting, the inputs of FNO omit physical variables, including rainfall and terrain DEM.
\textbf{PCNO without rainfall and terrain DEM:} In flood forecasting, PCNO receives inputs that exclude physical variables, including rainfall and terrain DEM.
 Training details for these baselines are available in Supplementary Tables 2-5. 
\section*{Data availability}
All datasets generated in this work are available online. The datasets for the KSE, Kolmogorov flow, and atmospheric modeling are available via the Zenodo repository at~\url{https://doi.org/10.5281/zenodo.17410273}. The datasets for the flood forecasting are available at~\url{https://doi.org/10.5281/zenodo.14017092}. 

\section*{Code availability}
The code implementing the baseline methods, PCNO, and DiffPCNO will be publicly accessible via the GitHub repository at~\url{https://github.com/HydroPML/PCNO}. Corresponding trained models will also be released upon the acceptance of this work.


\backmatter

\begin{appendices}






\end{appendices}


\bibliography{sn-bibliography}


\section*{Author contributions}
Q.X., J.B., and X.Z. conceived and designed the study. Q.X. conducted the data collection and generation, performed model training, and  analyzed the results. All authors contributed to detailed discussions of the results. Q.X. wrote the manuscript with input from all authors.

\section*{Competing interests}
The authors declare no competing interests.

\newpage
\setcounter{figure}{0}
\renewcommand{\thefigure}{\arabic{figure}}
\captionsetup[figure]{name=Extended Data Fig}

\begin{figure}[htp!]
	\centering
	{\includegraphics[width = 1.0\textwidth]{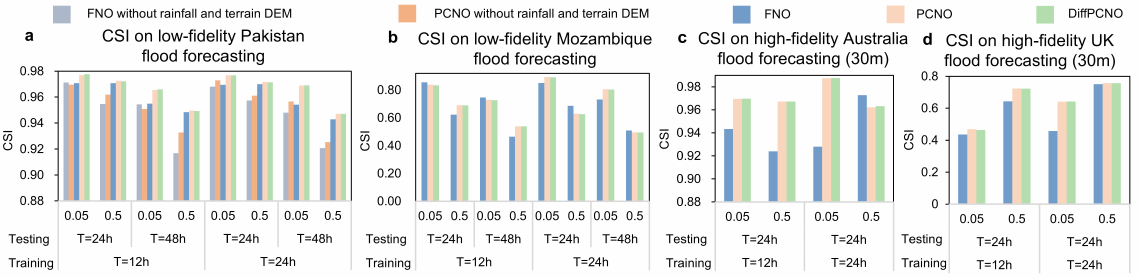}}
	\caption{CSI results of real-world flood inundation forecasting. 
\textbf{a-d} compare the CSI (higher values indicate better performance) at water depth thresholds of 0.05 m and 0.5 m for low-fidelity Pakistan flood forecasting, transferable Mozambique flood forecasting, high-fidelity Australia flood forecasting (30 m), and transferable UK flood forecasting (30 m). Comparisons of prediction performance at different flood depths reveal that, relative to FNO, PCNO with physical constraints and DiffPCNO with probabilistic learning consistently provide superior forecasts across varying water levels and time steps.}
	\label{fig:e1}
    \end{figure}

\begin{figure}[htp!]
	\centering
	{\includegraphics[width = 0.98\textwidth]{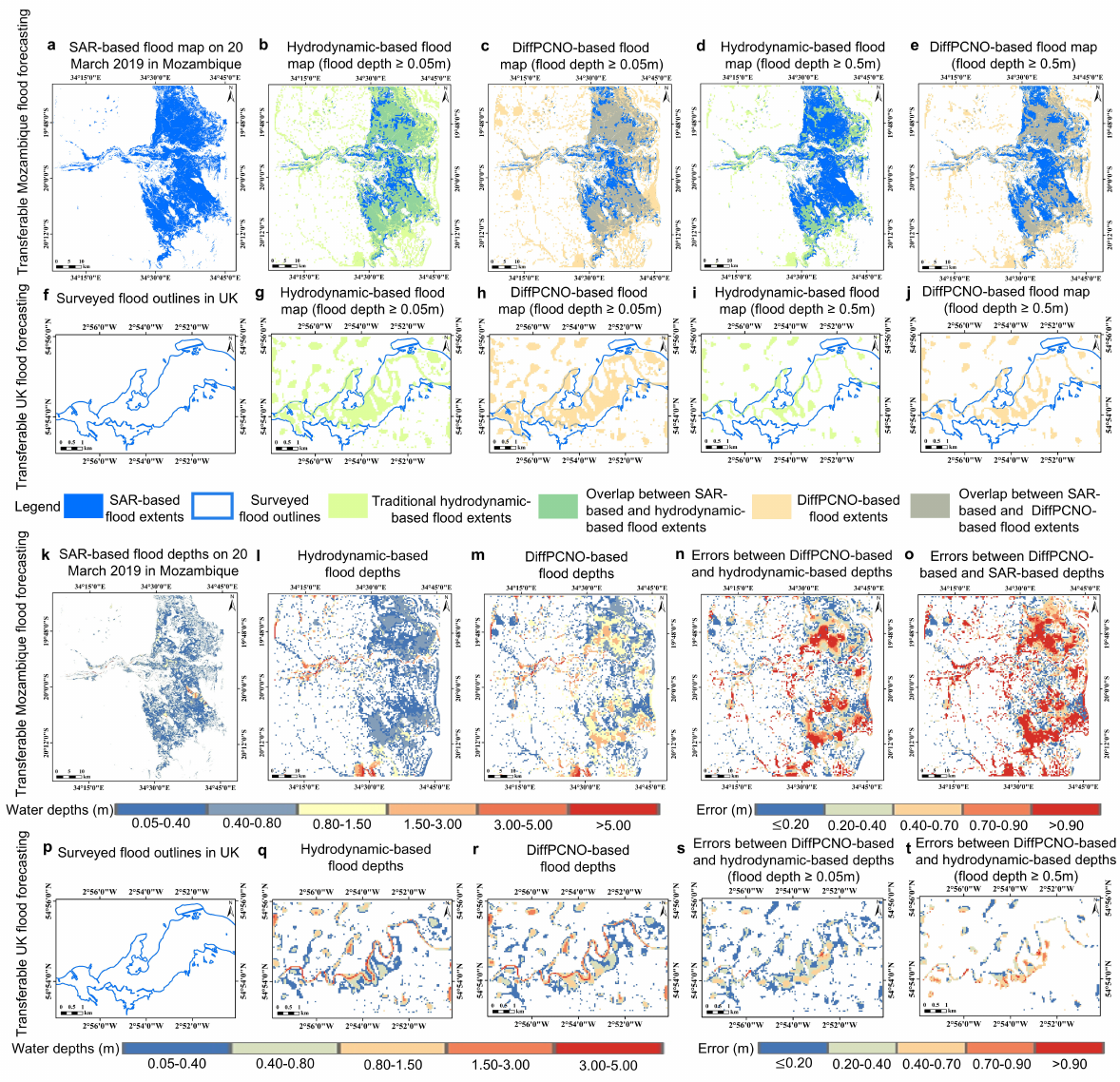}}
	\caption{Results of transferable flood inundation forecasting. Spatial variability of final flood inundation extents and depths across flood measurements (SAR and surveyed outlines), traditional hydrodynamic models, and DiffPCNO. \textbf{a,} SAR-based flood map on 20 March 2019 in Mozambique. \textbf{f,} Surveyed flood outlines in UK. 
    \textbf{b} and \textbf{g,}  Traditional hydrodynamic-based flood maps for Mozambique and UK (flood depth$\ge$0.05 m). \textbf{c} and \textbf{h,} DiffPCNO-based flood maps for Mozambique and UK (flood depth$\ge$0.05 m). 
    \textbf{d} and \textbf{i,}  Traditional hydrodynamic-based flood maps for Mozambique and UK (flood depth$\ge$0.5 m). \textbf{e} and \textbf{j,} DiffPCNO-based flood maps for Mozambique and UK (flood depth$\ge$0.5 m). \textbf{k,} SAR-based flood depths for Mozambique on 20 March 2019. SAR-based flood depths are extracted using the DEM and SAR-based flood extents~\cite{cohen2018estimating}. \textbf{p,} Surveyed flood outlines in UK. \textbf{l} and \textbf{q,} Traditional hydrodynamic-based flood depths for Mozambique and UK (flood depth$\ge$0.05 m). \textbf{m} and \textbf{r,} DiffPCNO-based flood depths for Mozambique and UK (flood depth$\ge$0.05 m).
    \textbf{n} and \textbf{s,} Spatial distribution of errors between DiffPCNO-based and hydrodynamic-based flood depths for Mozambique and UK (flood depth$\ge$0.05 m). \textbf{o,} Spatial distribution of errors between DiffPCNO-based and SAR-based flood depths for Mozambique.
    \textbf{t,} Spatial distribution of errors between DiffPCNO-based and hydrodynamic-based flood depths for UK (flood depth$\ge$0.5 m).
	}
	\label{fig:6}
\end{figure}

\appendix 
\renewcommand{\thesection}{\arabic{section}}
\setcounter{section}{0}

\newpage
\section*{Supplementary information}
\section*{Supplementary Notes}

\section{Kuramoto–Sivashinsky dynamics}
\subsection{Training data}
Data are generated on a 256-point spatial grid, with domain lengths $L$ sampled from $[0.9 \cdot 64,1.1 \cdot 64]$ and time steps $\Delta t$ drawn uniformly between 0.18 and 0.22 seconds. We follow the data generation setup of the work~\cite{brandstetter2022lie}, employing the method of lines with spatial derivatives computed via the pseudo-spectral approach. In this approach, derivatives are evaluated in the frequency domain by first applying a fast Fourier transform (FFT) to the data, multiplying by the corresponding frequency factors, and then transforming back to the spatial domain using the inverse FFT. Time integration is performed using an implicit Runge-Kutta scheme. For each trajectory in our dataset, the initial 360 solution steps are discarded and treated as a solver warmup.

The dataset is reproducible using the publicly available code~\cite{brandstetter2023LPSDA}. Following the settings of the work~\cite{lippe2023pde}, for 1D KSE with fixed viscosity, training data are generated by adjusting the repository command to produce 2048 training samples, with zero samples allocated for validation and testing. For validation and testing, we extend the rollout time and generate 128 samples each. For the 1D KSE with varying viscosity, we set the number of training samples to 4096, and 512 samples for validation and testing. Training trajectories consist of 140 time steps, whereas validation and test trajectories contain 400 time steps. In the  KSE with varying viscosity, the viscosity $\nu$ is incorporated by multiplying the fourth derivative estimate $\partial_x^4 u$ by $\nu$. For each training and test trajectory, $\nu$ is uniformly sampled from the interval [0.5, 1.5]. Data are initially generated in float64 precision and subsequently converted to float32 for storage and neural surrogate training. This conversion does not affect model training~\cite{lippe2023pde}.

\subsection{Task details}
We consider two scenarios. First, the standard KSE with fixed viscosity $\nu=1$, for which 2048 training and 128 test trajectories are generated. Second, the parameter-dependent KSE with varying viscosity $\nu$ sampled uniformly between 0.5 and 1.5, comprising 4096 training and 512 test trajectories.
 We evaluate the performance of PCNO (FNO), DiffPCNO, U-Net, PDE-Refiner, PDE-Refiner+, and the consistency model on the two-dimensional KSE.
 Each model is trained to predict the solution $u(t)$ given a single previous state $u(t-\Delta t)$. Long-term trajectories are generated recursively by feeding model predictions back as inputs for subsequent steps. Leveraging the capacity of neural networks to handle larger time steps, predictions are made at every fourth time step. Accordingly, each model predicts $u(t)$ from the preceding state $u(t- 4 \Delta t)$ together with the trajectory parameters $L$, $\Delta t$, and $\nu$. Following the setup~\cite{lippe2023pde}, PDE-Refiner and PDE-Refiner+ predict the residual between time steps $\Delta u(t)=u(t)-u(t-4 \Delta t)$, while other models directly predict $u(t)$. For the 1D KSE with fixed viscosity, $\Delta t$ and $\Delta x$ are used as the conditioning features for the surrogate models. In the case of 1D KSE with varying viscosity, $\Delta t, \Delta x$, and $\nu$ are employed as conditioning features.

\subsection{Training details}
Detailed hyperparameter specifications for the surrogate models are provided in Supplementary Table~\ref{table0}. All models are implemented in PyTorch and trained on a single NVIDIA RTX A6000 48 GB GPU.

\section{Kolmogorov turbulent flow}
\subsection{Training data}
For the two cases (velocity and vorticity forms) in the incompressible NS datasets, we generate a total of 1,200 samples using a pseudo-spectral Crank–Nicolson solver~\cite{li2020fourier}. The external forcing term is defined $f=0.1(\sin (2 \pi(x+y))+ \cos (2 \pi(x+y)))$. The system is defined over $x \in(0,1)^2$ and $t=1,2, \ldots, T$, initialized with a random vorticity field $w_0$ sampled from a Gaussian distribution, and subject to periodic boundary conditions. Experiments are conducted at a viscosity of $\nu=1 \times 10^{-3}$, with the final time $T$ progressively reduced as the dynamics become increasingly chaotic. The timestep is $\Delta t=10^{-4} \mathrm{~s}$, and the spatial resolution is fixed at $64 \times 64$ for training, validation, and testing. 

For the velocity form, datasets comprise 100, 300, 500, 800, and 1,000 training trajectories, along with 100 validation and 100 test trajectories. For the vorticity form, datasets comprise 1,000 training trajectories, 100 validation trajectories, and 100 test trajectories.

\subsection{Task details}
For the NS equations,  we project the ground truth vorticity field or velocity field from $T_{in}=10$ to each time step up to
$T>10$. Specifically, for the velocity form, the model predicts a rollout of $T=20$
  timesteps conditioned on the first 
 $T_{in}=10$ timesteps. For the vorticity form, the model predicts a rollout of $T=40$ subsequent timesteps conditioned on the first  $T_{in}=10$ timesteps. Compared with the velocity form, the vorticity formulation facilitates longer-term extrapolation.

\subsection{Training details}
The detailed training configurations of the surrogate models are provided in Supplementary Table~\ref{table02}. All models are implemented in PyTorch and trained on a single NVIDIA RTX A6000 GPU with 48 GB of memory.

\section{Real-world flood inundation forecasting}
\subsection{Training data}
Since flood water depth is generally much smaller than its horizontal extent, flow dynamics can be approximated by the depth-averaged 2D  shallow water equations (SWE)~\cite{de2012improving,xu2024large}. By neglecting the convective acceleration term, the SWE for flood modeling  can be written as,
\begin{equation}
\begin{gathered}
\frac{\partial h}{\partial t}+\frac{\partial q_x}{\partial x}+\frac{\partial q_y}{\partial y}=R-I, \\
\frac{\partial q_x}{\partial t}+g h \frac{\partial(h+z)}{\partial x}+\frac{g n^2\left|q\right| q_x}{h^{7 / 3}}=0, \\
\frac{\partial q_y}{\partial t}+g h \frac{\partial(h+z)}{\partial y}+\frac{g n^2\left|q\right| q_y}{h^{7 / 3}}=0,
\end{gathered}
\label{eq44}
\end{equation}
where $h$ is the water height that we will predict, relative to the terrain elevation $z$. 
$t$ is the time index. $x, y$ are the spatial horizontal coordinates. 
$q = (q_{x}, q_{y})$ is the discharge per unit width. $R$ represents the rainfall rate, and $I$ is the infiltration rate. 
$n$ is Manning's friction coefficient.  

Training data are generated by discretizing Eq.~\ref{eq44} from its continuous form to a discrete domain using a finite difference (FD) method, a widely accepted hydrodynamic technique for simulating flood scenarios. 
\subsubsection{FloodCastBench}
FloodCastBench dataset~\cite{xu2025floodcastbench} includes four large-scale flood events in supporting ML models for spatiotemporal flood forecasting, cross-regional, and downscaled flood forecasting. The locations of these events are presented in Supplementary Fig.~\ref{fig:s1}. These events are as follows:
\begin{itemize}
\item 	Pakistan Flood (2022).
During the summer monsoon season of 2022, Pakistan suffered catastrophic flooding induced by exceptionally intense and prolonged rainfall, ranking among the most devastating natural disasters in the nation’s recorded history. The event affected nearly one-third of the country’s population, displacing approximately 32 million individuals and resulting in 1,486 fatalities, including 530 children. The total economic loss was estimated to exceed $\$30$ billion~\cite{Bhutto2022}. Beyond the immediate consequences, the widespread destruction of agricultural fields has raised concerns of potential famine, and there is a looming threat of disease outbreaks in temporary shelters~\cite{nanditha2023pakistan}.
The study area encompasses the regions most severely impacted, including the southern provinces of Balochistan, Sindh, and Punjab, covering an approximate total area of 85,616.5 $km^{2}$. The Indus River Basin, a key hydrological system, plays a pivotal role in governing the region’s drainage and flood dynamics. Between August 18 and August 31, 2022, a substantial expansion in flood extent was observed across the study area~\cite{xu2024large}. Accordingly, flood simulations are conducted over this 14-day period to capture the hydrodynamic evolution of the event.

\item   Mozambique Flood (2019).
In March 2019, Tropical Cyclone Idai made landfall near the coastal city of Beira in Sofala Province, central Mozambique, bringing torrential rainfall and strong winds that persisted for more than a week. As a result, major rivers including the Pungwe and Buzi overflowed, inundating extensive low-lying regions~\cite{guo2021mozambique}. The disaster caused widespread devastation, with approximately 4,000 homes rendered damaged or uninhabitable, 1,600 injuries, and 603 fatalities reported across Mozambique, Zimbabwe, and Malawi~\cite{undp2019}. 
The study area, centered around Beira, encompasses approximately 6,190.9 $km^{2}$. Flood simulations cover the period from March 14 to March 20, 2019, based on recorded precipitation data and SAR-derived flood extents.

\item   Australia Flood (2022). 
In early 2022, eastern Australia experienced extensive flooding initiated in February and subsequently intensified following additional heavy rainfall episodes. Severe inundations were reported across Queensland and New South Wales, prompting large-scale evacuations and significant damage to residential properties and infrastructure. The Richmond River Basin, serving as a critical drainage network, exerts a substantial influence on the hydrological regime of the study region~\cite{Australia_2022}.
The selected area, encompassing parts of Ballina, covers a total of approximately 1,361.3 $km^{2}$. The simulation period extends from February 20 to March 2, 2022, utilizing rainfall datasets and SAR-derived flood observations to reconstruct flood dynamics.

\item	UK Flood (2015). 
The study area representing the UK flood event is located in northwest England and is dominated by the 145 km long River Eden, which flows from the southeast to the northwest. The catchment consists of four main tributaries—Caldew, Petteril, Eamont, and Irthing—and is characterized by steep upstream topography, which contributes to frequent fluvial flooding and demands prompt flood management responses. The downstream city of Carlisle has historically experienced recurrent flood disasters, with the December 2015 event being particularly severe, causing extensive damage~\cite{ming2020real}.
The designated study region spans approximately 135.5 $km^{2}$. The flood was primarily triggered by extreme rainfall between December 4 and 7, 2015, lasting three consecutive days.
\end{itemize}

\subsubsection{Data requirements for FloodCastBench}
The data required for conventional numerical methods, such as a finite difference (FD) scheme, encompass topographical data, land cover maps, real-time gridded rainfall data, and  flood measurement data within the study area.

Specifically, a high-resolution (30 m) forest and buildings removed Copernicus digital elevation model (FABDEM)~\cite{hawker202230} from COPDEM30 is utilized for flood simulation.

Land cover information is useful for estimating and adjusting friction (Manning coefficient).  Land cover information in the study area can be subtracted from the Sentinel-2 land use/land cover dataset~\cite{karra2021global}. It is produced by a deep learning model by classifying Sentinel-2 data into 9 classes, available at a spatial resolution of up to 10 m for the study area.

The rainfall data is a grid-based data set at $0.1^{\circ} \times 0.1^{\circ}$ spatial resolution and half-hourly temporal resolution from GPM-IMERG. Utilizing bilinear interpolation to resample the data, a grid-based rainfall data at   5-minute temporal resolution and  30 m $\times$ 30 m spatial resolution is obtained.

Flood measurement data comprise SAR-based  flood maps and surveyed flood outlines, which are used for the calibration and verification of hydrodynamic-based flood inundation. Specifically, SAR-based flood maps for Mozambique (acquired on 20 March 2019) and Australia (acquired on 2 March 2022) are obtained from the UrbanSARFloods dataset~\cite{zhao2024urbansarfloods}, while the SAR-based flood map for Pakistan (acquired on 30 August 2022) is sourced from the Global Flood Awareness System (GloFAS)~\cite{salamon_wagner_2022}. All SAR-based flood maps from both UrbanSARFloods and GloFAS have a spatial resolution of 20 m. In addition, surveyed flood outlines for the 2015 UK flood event are publicly available under the UK Open Government Licence and are accessible online~\cite{UK_2015}.

\subsubsection{Implementation details of numerical methods}
The inputs to the hydraulic model include an elevation map, initial conditions, boundary conditions, and rainfall forcing. For the 2022 Pakistan flood event, the initial conditions are specified using water depths retrieved from SAR observations, whereas for other events, they are obtained by prerunning a finite-difference solver over a dry domain. The boundary conditions for the Pakistan case consist of an inflow boundary along the Indus River and a free outflow boundary defined with a valley slope of 0.2. The inflow discharge from 18 to 31 August is prescribed using hydrological station records. For the remaining flood events, only the free outflow boundary with a valley slope of 0.2 is applied. Rainfall is provided as a spatially distributed field, and Manning’s roughness coefficients are assigned according to land-cover classifications. Further details are available in the FloodCastBench dataset~\cite{xu2025floodcastbench}.

We implement the numerical solution using Python. The temporal resolution is selected to satisfy the Courant–Friedrichs–Lewy (CFL) stability criterion for hyperbolic systems, ensuring numerical convergence and physical fidelity.
Given the fine temporal discretization (less than 10 s), the 14-day simulation of the 2022 Pakistan flood at 480m $\times$ 480m resolution is executed on an NVIDIA A6000 GPU and completes in approximately one week. The 3-day simulation of the 2015 UK flood at 30m $\times$ 30m resolution requires approximately 4 h, while the 10-day simulation of the 2022 Australia flood at the same resolution completes in around two days. The 4-day simulation of the 2019 Mozambique flood at 480m $\times$ 480m resolution on the same GPU requires approximately 16 h.

Finally, we construct the FloodCastBench dataset for both low-fidelity and high-fidelity flood forecasting, as illustrated in Supplementary Fig.~\ref{fig:s02}. Due to the substantial variability in spatiotemporal dynamics, hydrometeorological drivers, elevation DEM, and land use and land cover across flood events, accurate and reliable spatiotemporal flood forecasting remains a significant challenge for ML-based surrogate models. We therefore conduct a comprehensive evaluation of surrogate model performance in different regimes: (1) spatiotemporal forecasting under distinct rainfall forcings, such as the testing datasets in the 2022 low-fidelity Pakistan flood forecasting  and 2022 high-fidelity Australia flood forecasting; (2) downscaled and transferable forecasting across different scenarios, such as the  datasets in the 2019 Mozambique flood and the 2015 UK flood.

\subsection{Task details}
In all tasks, we forecast flood depth from time $t=1$ to $T$.
Low-fidelity forecasting utilizes the Pakistan and Mozambique flood datasets, which have a spatial resolution of 480m and a temporal resolution of 300 seconds. Two long-term flood forecasting experiments for the Pakistan flood are performed: training samples of $T=144$ (12 12 hours) and testing at $T=288$ (24 hours) or $T=576$ (48 hours); training samples of $T=288$ (24 hours) and testing at $T=288$ (24 hours) or $T=576$ (48 hours). The Pakistan flood dataset covers a 14-day period, with the first 10 days used for training, the 11th and 12th days for validation, and the final 2 days for testing. For cross-regional forecasting, the Mozambique flood dataset spans 4 days, yielding 4 samples at $T=288$ and 2 samples at $T=576$.

High-fidelity forecasting utilizes the Australia and UK flood datasets, which have spatial resolutions of 60 m or 30 m, and a common temporal resolution of 300 seconds.
 For high-fidelity forecasting, we train the model on the Australia flood (60 m) and evaluate transferability on the UK flood. Two long-term setups are conducted: training samples of $T=144$ (12 hours) and testing at $T=288$ (24 hours); and training samples of $T=288$ (24 hours) and testing at $T=288$ (24 hours). The Australia dataset encompasses a ten-day period, of which the first eight days are allocated for model training, the ninth day is used for validation, and the tenth day serves as the test set. The UK flood dataset, spanning three days, is utilized for cross-regional forecasting, yielding 3 samples at $T=288$. For the downscaling (zero-shot super resolution) experiments, the model is trained on 60m data and directly evaluated on the 30m Australia test set and the UK dataset.
 
\subsection{Training details}
The surrogate models are formulated using input data comprising 2D spatial coordinates, temporal information, initial water depth, Digital Elevation Model (DEM), and rainfall forcing over the prediction horizon. The detailed training configurations of the surrogate models are provided in Supplementary Table~\ref{table04}. All models are implemented in PyTorch and are trained on a single NVIDIA RTX A6000 GPU with 48 GB of memory. Supplementary Table~\ref{table06} presents the runtime of the surrogate models and the hydrodynamic method used for flood inundation forecasting. A runtime comparison of flood events predicted by the surrogate models and the hydrodynamic method over the same time period further demonstrates the computational efficiency of the surrogate models.

\section{Atmospheric modeling}
\subsection{Training  data} 
In the approximation for negligible depth compared to horizontal scales, the shallow water equations for simulating gravity waves in the Earth’s atmosphere are derived from the conservation of mass and momentum,
\begin{equation}
\begin{aligned}
& \frac{\partial \eta}{\partial t}+\nabla \cdot(\mathbf{u} h) =0, \quad
\frac{\partial \zeta}{\partial t}+\nabla \cdot(\mathbf{u}(\zeta+f))  =0, \\
& \frac{\partial \mathcal{D}}{\partial t}-\nabla \times(\mathbf{u}(\zeta+f)) =-\nabla^2\left(\frac{1}{2}|\mathbf{u}|^2+g \eta\right).
\end{aligned}
\end{equation}
The equations are solved in spherical coordinates, with latitude $\theta \in [-\pi, \pi]$ and longitude $\lambda \in [0, 2\pi]$, on a sphere of radius $R = 6371 \mathrm{~km}$. The relative vorticity is $\zeta = \nabla \times \mathbf{u}$, and the divergence is $\mathcal{D} = \nabla \cdot \mathbf{u}$, where $\mathbf{u}$ denotes the velocity. $\eta$ represents the displacement from the rest height $H = 8500 \mathrm{~m}$. The layer thickness is given by $h = \eta + H - H_0$, where $H_0 = H_0(\lambda, \phi)$ represents Earth’s orography, and $\phi$ denote the colatitude. The gravitational acceleration is $g = 9.81\mathrm{~ms^{-2}}$, and the Coriolis parameter is $f = 2 \Omega \sin \theta$, with $\Omega = 7.29 \times 10^{-5} \mathrm{~s^{-1}}$. The simulations are initialized from a state of rest ($\mathbf{u} = 0$), with random perturbations in $\eta$ characterized by wavelengths of approximately 2000-4000~km and maximum amplitudes of 2000 m. These initial perturbations propagate globally as gravity waves with a phase speed $c_{\mathrm{ph}} = \sqrt{g h}$ (approximately $300 \mathrm{~ms^{-1}}$), undergoing nonlinear interactions with one another and with Earth’s orography. 

We employ the general circulation model SpeedyWeather.jl to solve the shallow water equations. The initial velocity $\mathbf{u}_0$ and surface displacement $\eta_0$ on the computational grid are obtained via the spherical harmonic transform as,
\begin{equation}
\mathbf{u}_0=0, \quad
\eta_0 =A \sum_{\ell=0}^{\ell_{\max }} \sum_{m=-\ell}^{\ell} \eta_{\ell, m} Y_{\ell, m},
\end{equation}
where the amplitude $A$ is chosen such that $\left(\left|\eta_0\right|\right)=2000 \mathrm{~m}$. The spherical harmonics are represented as $Y_{\ell, m}$, with degree $\ell \geq 0$ and order $m$ with $-\ell \leq m \leq \ell$. Random coefficients $\eta_{\ell, m}$ are assigned using a standard complex normal distribution $\mathcal{C} \mathcal{N}(0,1)=\mathcal{N}\left(0, \frac{1}{2}\right)+i \mathcal{N}\left(0, \frac{1}{2}\right)$ for degrees $10 \leq \ell<20$. For the zonal modes ($m=0$), coefficients are drawn from a real normal distribution $\eta_{\ell, 0} \sim \mathcal{N}(0,1) $, while all other coefficients are set to zero. The corresponding wavelengths are given by $2 \pi R / \ell$, ranging approximately from 2000 to 4000 km.
The simulation resolution is set by the maximum resolved spherical harmonic degree, $\ell_{\max }=63$, corresponding to the commonly used T63 spectral truncation in numerical weather prediction. This spectral resolution is combined with a regular longitude-latitude grid of $192 \times 95$ points ($\Delta \lambda=\Delta \theta=1.875^{\circ}$, approximately 200 km at the equator, with no points at the poles), also referred to as a full Clenshaw-Curtis grid~\cite{hotta2018nestable}. Non-linear terms are evaluated in physical space, while linear terms are computed in spectral space, with transformations between the two performed at each time step. This hybrid approach is widely employed in global numerical weather prediction models. For numerical stability, a horizontally implicit diffusion term of the form $-\nu \nabla^8 \zeta$ and $-\nu \nabla^8 \mathcal{D}$ is added to the vorticity and divergence equations, respectively. The fourth-power Laplacian is highly scale-selective, removing energy only at the highest wavenumbers while leaving the larger-scale flow largely unaffected. 

Time integration in SpeedyWeather.jl employs a RAW-filtered Leapfrog scheme~\cite{williams2011raw} with a time step $\Delta t=15 \mathrm{~min}$ at T63 resolution. At this time step, the CFL number $C=c_{p h} \Delta t(\Delta x)^{-1}$ with equatorial $\Delta x=2 \pi R \frac{\Delta \lambda}{360^{\circ}}$, typically ranges between 1 and 1.4 for phase speeds $c_{p h}=\sqrt{g h}$ between of 280-320 $\mathrm{~ms}^{-1}$. By employing a centred semi-implicit Leapfrog integration, the simulation maintains stability without the need for excessive damping of gravity waves through larger time steps or a fully implicit backward scheme. Following the simulation setups~\cite{liu2024harnessing}, we establish a physically well-defined dataset for investigating non-linear gravity-wave propagation. A total of 1,200 samples are generated in atmospheric modeling, partitioned into 1,000/100/100 for training, validation, and testing, respectively.
\subsection{Task details}
 All models generate rollouts of $T = 14$ time steps conditioned on the first step ($T_{in} = 1$). For the 2D model predicting atmospheric variables, we use spatial coordinates and initial conditions as inputs to forecast the layer thickness $h$, zonal wind velocity $u_x$, and meridional wind velocity $u_y$.

To ensure divergence-free predictions, both the 2D and 3D models are trained to learn the transformed variables $(u_x h, u_y h \sin \theta, R h \sin \theta)$. Specifically, mass conservation in spherical coordinates is expressed as,
\begin{equation}
\frac{\partial \eta}{\partial t}+\nabla \cdot(\mathbf{u} h)=0,
\end{equation}
where $\eta$ denotes the displacement from the atmospheric rest height $H=8500 \mathrm{~m}, h=\eta+H- H_0$ represents the layer thickness, and $H_0=H_0(\lambda, \phi)$ corresponds to the Earth's orography, with $\lambda$ and $\phi$ denoting longitude and colatitude, respectively. We note that $\partial \eta / \partial t=\partial h / \partial t$, and the divergence term in spherical coordinates can be rewritten as:
\begin{equation}
\nabla \cdot(\mathbf{u} h)=\frac{1}{R \sin \theta} \frac{\partial\left(u_x h\right)}{\partial \lambda}+\frac{1}{R \sin \theta} \frac{\partial\left(u_y h \sin \theta\right)}{\partial \theta}.
\end{equation}
Here we use latitude instead of colatitude. Consequently, the mass conservation equation can be expressed as,
\begin{equation}
\begin{aligned}
\frac{1}{R \sin \theta} \frac{\partial\left(u_x h\right)}{\partial \lambda}+\frac{1}{R \sin \theta} \frac{\partial\left(u_y h \sin \theta\right)}{\partial \theta}+\frac{\partial \eta}{\partial t} & = \\
\frac{\partial\left(u_x h\right)}{\partial \lambda}+\frac{\partial\left(u_y h \sin \theta\right)}{\partial \theta}+R \sin \theta \frac{\partial \eta}{\partial t}&= \\
\frac{\partial\left(u_x h\right)}{\partial \lambda}+\frac{\partial\left(u_y h \sin \theta\right)}{\partial \theta}+\frac{\partial(R h \sin \theta)}{\partial t} &= 0.
\end{aligned}
\end{equation}
Thus, we configure the model to learn the variables $(u_x h, u_y h \sin \theta, R h \sin \theta)$, ensuring that the predicted output field is divergence-free. Once the output $\mathbf{u}=\left(u_x h, u_y h \sin \theta, R h \sin \theta\right)$ is obtained, we post-process it to recover the physical variables $\left(u_x, u_y, h\right)$. For instance, $u_y$ is derived by dividing the second component $u_y h \sin \theta$ by the third component $R h \sin \theta$ and multiplying the result by $R$. This formulation allows the direct application of the proposed mass-conserving projection (Supplementary Notes 7.2) for 3D models.

\subsection{Training  details} 
Comprehensive hyperparameter details for both the 2D and 3D models are provided in Supplementary Table~\ref{table2}. All models are implemented in PyTorch and trained on a single NVIDIA RTX A6000 48 GB GPU. Supplementary Table~\ref{table3} reports the runtime of the 2D models for atmospheric modeling. Notably, DiffPCNO requires only 0.629 seconds to generate a single prediction, reflecting the efficiency of the consistency model’s sampling procedure.


\section{Noether’s theorem}
\subsection{Introduction of Noether’s theorem}
Conservation laws describe the physical properties of the spatiotemporal dynamics systems and are widely used to detect integrability, linearization, and enhance the accuracy of numerical solutions. Notably, Noether's theorem established a significant connection between symmetries and conservation laws for physical systems~\cite{noether1971invariant, bluman2010applications}.

To elucidate  this connection, consider the nonlinear function $F(\mathbf{u}_{t})$, with $\mathbf{u}_0(\mathbf{x})$ as the initial condition and $\mathbf{u}_{\mathbf{t}}(\mathbf{x})$ as the resulting dynamical response. Translational invariance implies that translating the input function $\mathbf{u}$ before applying the nonlinear function $F$ yields the same result as applying $F$ directly. As such, the resultant physical model remains invariant across spatial locations, and Noether's theorem guarantees the conservation of linear momentum. 
Furthermore, rotational invariance means that rotating the input function $\mathbf{u}$ first and then applying the nonlinear function $F$ will lead to the same result as applying $F$ directly. As such, the described physical model remains invariant under rotations against the origin, and Noether's theorem guarantees the conservation of angular momentum.

\subsection{Symmetries lead conservation laws}
\label{A.2}
Noether’s theorem asserts that for every symmetry in a physical system, there exists a corresponding conservation law~\cite{kara2002basis,arnol2013mathematical}. This principle is universally applicable to all physical systems. Nevertheless, to quantitatively define a symmetry in relation to Noether’s theorem, or to identify the specific conserved quantities resulting from these symmetries, it is typically necessary to prove the theorem within the framework of Lagrangian systems.

Mathematically, assume that the spatiotemporal dynamical system is conservative, with the Lagrangian form corresponding to the nonlinear function  $F$ denoted as $L(\mathbf{x}, \mathbf{u})$. $\mathbf{x}$ and $\mathbf{u}=\frac{\mathbf{x}}{d t}$ represent the generalized coordinates and velocity.   for every symmetry in a physical system, meaning a transformation that changes the physical system only by $\delta L=\frac{d F_{L}}{d t}$, there will be an associated conserved quantity of the form,
\begin{equation}
Q=\sum_i p_i \delta \mathbf{x}_i-F_{L},
\label{20}
\end{equation}
 where $p_i=\frac{\partial L}{\partial \mathbf{u}_i}$
 denotes the generalized momentum.  $\delta \mathbf{x}_i$
 represents changes in the generalized coordinates of the system resulting from the given symmetry transformation, and $F_{L}$
 is a function where the change in the Lagrangian corresponds to a total time derivative under the symmetry transformation, often simply zero.

\textbf{Proof.}   The variation in the Lagrangian $L(\mathbf{x}, \mathbf{u})$ is given by the chain rule as,
 \begin{equation}
 \begin{aligned}
\delta L&=\sum_i\left(\frac{\partial L}{\partial \mathbf{x}_i} \delta \mathbf{x}_i+\frac{\partial L}{\partial \mathbf{u}_i} \delta \mathbf{u}_i\right) \\
&=\sum_i\left(\dot{p}_i \delta \mathbf{x}_i+p_i \delta \dot{\mathbf{x}}_i\right) \\
&=\sum_i \frac{d}{d t}\left(p_i \delta \mathbf{x}_i\right)=\frac{d}{d t}\left(\sum_i p_i \delta \mathbf{x}_i\right),
\end{aligned}
\label{21}
\end{equation}
where the time derivative of the generalized momentum $\dot{p}_i=\frac{\partial L}{\partial \mathbf{x}_i}$ is given by the Euler-Lagrange equation.

Substituting Eq.~\ref{21} into $\delta L=\frac{d F_{L}}{d t}$  and rearranging terms, we obtain,
\begin{equation}
\begin{aligned}
 \frac{d}{d t}\left(\sum_i p_i \delta \mathbf{x}_i\right)=\frac{d F_{L}}{d t}  \Rightarrow \frac{d}{d t}\left(\sum_i p_i \delta \mathbf{x}_i-F_{L}\right)=0.
\end{aligned}
\end{equation}
Thus, the quantity  $Q$ within the parentheses (Eq.~\ref{20}) is identified as a conserved quantity.

 A translation in this generalized coordinate $ \mathbf{x}_i$ simply means that we shift the coordinate by an infinitesimal amount $\delta  \mathbf{x}_i$, denoted as $\mathbf{x}_i \rightarrow \mathbf{x}_i+\delta \mathbf{x}_i$. Since the translation $\delta \mathbf{x}_i$
 is constant, the time derivative of the coordinate   remains unchanged. Consequently, the variation in the Lagrangian is expressed as $\delta L=\sum_i\left(\frac{\partial L}{\partial \mathbf{x}_i} \delta \mathbf{x}_i+\frac{\partial L}{\partial \mathbf{u}_i} \delta \mathbf{u}_i\right)=\sum_i\frac{\partial L}{\partial \mathbf{x}_i} \delta \mathbf{x}_i$. For this transformation to be a symmetry, we require $\delta L=0$, which implies that the Lagrangian is independent of the specific coordinate $\mathbf{x}_i$
 undergoing the transformation ($\frac{\partial L}{\partial \mathbf{x}_i}=0$). Thus, the independence of the Lagrangian with respect to a given coordinate signifies that translations of this coordinate are symmetries of the physical system.

 According to Eq.~\ref{20}, we find a conserved quantity of the form ($F_{L}=0$),
\begin{equation}
Q=\sum_i p_i \delta \mathbf{x}_i-F_{L}=\sum_i p_i \delta \mathbf{x}_i,
\end{equation}
where the $\delta \mathbf{x}_i$
 are constant, the conserved quantity here is just the 
$p_i$ (the generalized momentum). Thus, we have established that translation symmetry in a given generalized coordinate leads to the conservation of generalized momentum. Specifically, if this coordinate is a spatial coordinate (translation), the associated conserved generalized momentum is linear momentum. Conversely, if this coordinate is an angular coordinate (rotation), the conserved generalized momentum would be angular momentum.

\section{Momentum-conserving projection}
\subsection{Translation invariance}
To establish translation invariance in momentum-conserving projection, we employ the correlation theorem~\cite{cohen2016group}, which gives that,
\begin{equation}
\begin{aligned}
& [L_g[\mathcal{G}] \star W](x)
 = \mathcal{F}^{-1}\left(\mathcal{F}\left(L_g[\mathcal{G}]\right) \cdot \mathcal{F}\left(W\right)\right) \\
& =\sum_y \mathcal{G}(y-g) W(y-x) =\sum_y \mathcal{G}(y) W(y-(x-g))\\
& =\left[\mathcal{G} \star W]\right(x-g) = \mathcal{F}^{-1}\left(\mathcal{F}\left(\mathcal{G}\right) \cdot \mathcal{F}\left(W\right)\right),
\label{eq7}
\end{aligned}
\end{equation}
where $L_g[\mathcal{G}]$ is a translation on the
output of the surrogate model $\mathcal{G}$, defined as $L_g[\mathcal{G}] =\tilde{\mathcal{G}}$ with $\tilde{\mathcal{G}}(\mathbf{x}+g)=\mathcal{G}(\mathbf{x})$.
$g \in \mathbb{R}^d$ is a translation vector defined in the output space dimensions. Hence, the Fourier layer serves as a translation-invariant projection, inherently preserving linear momentum without the need to impose physical constraints explicitly.

\subsection{Rotational invariance}
To establish rotational invariance,  leveraging the connection of transformations between the frequency and physical domains, our momentum-conserving projection is derived as,
\begin{equation}
\begin{aligned}
& [L_R[\mathcal{G}] \star W](x) =\sum_y \mathcal{G}( R^{-1}y) W(y-x) \\ &=\sum_y \mathcal{G}(y)\left(W(R(y-x))\right)  =\sum_y \mathcal{G}(y)\left(L_R[W])\right) =\mathcal{F}^{-1}\left(\left(L_R[\mathcal{F}(W)]\right) \cdot \mathcal{F}(\mathcal{G}) \right),
\label{eq8}
\end{aligned}
\end{equation}
where $L_R[\mathcal{G}]$ is a rotation on the
output of the surrogate model $\mathcal{G}$, defined as $L_R[\mathcal{G}] = \tilde{\mathcal{G}}$ with $\tilde{\mathcal{G}}(R\mathbf{x})=R\mathcal{G}(\mathbf{x})$. Here $R \in \mathbb{R}^{d_m \times d_m}$ is an orthogonal matrix, where $d_m$ denotes the dimension of the matrix. Eq.~\ref{eq8} implies that, to achieve rotation invariance in the output space, the projection operator should realize a rotation-invariant convolution in the frequency domain by transforming 
$\mathcal{F}(W)$ with a designed rotation-invariant kernel. Following the approaches~\cite{helwig2023group, xu2024physics}, such a kernel can be constructed by parameterizing only half of the complex-valued weights along a designated axis and generating the remaining half through a symmetric rotation about the kernel's center. This design enforces rotational invariance in the learned representations.
\subsection{Implementation details of the momentum-conserving projection}
To ensure that the kernel function $W$ is real valued and that correlation and convolution are equivalent in Eq.~\ref{eq7} and Eq.~\ref{eq8}, the rotation-invariant kernel $L_R[\mathcal{F}(W)]$ needs to be Hermitian. That is, $L_R[\mathcal{F}(W)]=L_R[\mathcal{F}(W)]^*$, where $L_R[\mathcal{F}(W)]^*$ denotes complex conjugation of $L_R[\mathcal{F}(W)]$.
To impose this constraint, the two terms corresponding to the upper and lower triangles of the rotation-invariant kernel are conjugates of each other.

To achieve rotational invariance,  it is crucial that the modes obtained from the FFT $\mathcal{F}(\mathcal{G}\left[\mathbf{u}_t; \theta\right])$ align correctly with  $L_R[\mathcal{F}(W)]$. While $L_R[\mathcal{F}(W)]$ has a centered origin, the FFT of $\mathcal{G}\left[\mathbf{u}_t; \theta\right]$ typically does not ensure this. Thus, a frequency shift is applied to center the zero-frequency component following the FFT. After the rotation-invariant convolution in the frequency domain, the frequency shift is reversed, and the inverse FFT is performed.
Notably, 
padding is utilized to mitigate potential numerical artifacts arising from the periodic constraints of the Fourier transform when addressing non-periodic problems. 

\subsection{Translation- and rotation-invariant properties of the momentum-conserving projection}
\label{A.3}

Herein, we present a rigorous proof of the translation and rotation invariance properties of the momentum-conserving projection. Formally,
\begin{equation}
\begin{aligned}
&\mathcal{D}_{mom}(\tilde{\mathcal{G}})=W_{inv} \tilde{\mathcal{G}}+W_{inv}\left(\mathcal{F}^{-1}\left(\left(L_R[\mathcal{F}(W)]\right) \cdot \mathcal{F}(\tilde{\mathcal{G}}) \right)\right), \\
&=W_{inv} \mathcal{G}+W_{inv}\left(\mathcal{F}^{-1}\left(\left(L_R[\mathcal{F}(W)]\right) \cdot \mathcal{F}(\mathcal{G}) \right)\right) = \mathcal{D}_{mom}(\mathcal{G}), \\
&\tilde{\mathcal{G}}\left(R\mathbf{x}+g\right) =\left\{\begin{array}{l}
\tilde{\mathcal{G}}\left(Rx+g\right)=R\mathcal{G}\left(x\right), \mathbf{x} \text{ is scalar};  \\
\tilde{\mathcal{G}}\left(Rx+g, Ry+g\right)=R\mathcal{G}\left(x,y\right), \mathbf{x} \text{ is vector}.
\end{array}\right.
\end{aligned}
\label{eq.24}
\end{equation}

\textbf{Proof.}  From the translation invariance in Eq.~\ref{eq7} and the rotation invariance in Eq.~\ref{eq8}, it can be deduced that,
\begin{equation}
\begin{aligned}
\mathcal{F}^{-1}\left(\left(L_R[\mathcal{F}(W)]\right) \cdot \mathcal{F}(\tilde{\mathcal{G}}) \right) = \mathcal{F}^{-1}\left(\left(L_R[\mathcal{F}(W)]\right) \cdot \mathcal{F}(\mathcal{G}) \right).
\end{aligned}
\end{equation}
Thus, after applying the invariant convolutions $W_{inv}$ with rotational symmetry, we can derive,
\begin{equation}
\begin{aligned}
W_{inv}\left(\mathcal{F}^{-1}\left(\left(L_R[\mathcal{F}(W)]\right) \cdot \mathcal{F}(\tilde{\mathcal{G}}) \right)\right) = W_{inv}\left(\mathcal{F}^{-1}\left(\left(L_R[\mathcal{F}(W)]\right) \cdot \mathcal{F}(\mathcal{G}) \right)\right).
\end{aligned}
\label{eq.26}
\end{equation}
Furthermore, based on the translation and rotation invariance properties of the invariant convolutions $W_{inv}$ with rotational symmetry~\cite{cohen2016group}, we can deduce that,
\begin{equation}
\begin{aligned}
W_{inv} \tilde{\mathcal{G}} = W_{inv} \mathcal{G}.
\end{aligned}
\label{eq.27}
\end{equation}
By integrating Eq.~\ref{eq.26} with Eq.~\ref{eq.27}, we can establish Eq.~\ref{eq.24} ($\mathcal{D}_{mom}=\mathcal{D}_{mom}(\mathcal{G})$). Thus, the translation- and rotation-invariance properties of the momentum-conserving projection are conclusively proven.
By integrating the relationship between momentum conservation laws and symmetries established by Noether’s theorem, we theoretically demonstrate the effectiveness of the momentum-conserving projection layer in preserving momentum for spatiotemporal dynamics.
\section{Mass-conserving projection}
\subsection{Mass-conserving projection in the time-independent systems}
The mass-conserving projection in the time-independent systems requires that the divergence of the field vanish,
\begin{equation}
\mathcal{F}(\frac{\partial \mathbf{u}_1}{\partial \mathbf{x}_1}+\frac{\partial \mathbf{u}_2}{\partial \mathbf{x}_2})=0 \quad \Longrightarrow \quad k_1 \tilde{\mathbf{u}}_1(k)+k_2 \tilde{\mathbf{u}}_2(k)=0.
\end{equation}
Applying the Fourier transform $\mathcal{F}$, and using the first-order gradient solution $\mathcal{F}\left(\partial f / \partial x_j\right)=i k_j \tilde{f}(k)$, the divergence-free condition becomes,
\begin{equation}
\mathcal{F}\left(\frac{\partial \mathbf{u}_1}{\partial x_1}+\frac{\partial \mathbf{u}_2}{\partial x_2}\right)=i k_1 \tilde{\mathbf{u}}_1(k)+i k_2 \tilde{\mathbf{u}}_2(k)=0 \quad \Longleftrightarrow \quad k_1 \tilde{\mathbf{u}}_1(k)+k_2 \tilde{\mathbf{u}}_2(k)=0 .
\end{equation}

According to the discrete Helmholtz decomposition~\cite{girault2012finite}, an arbitrary vector field $\tilde{\mathbf{v}}(k)$ in Fourier space can be uniquely decomposed into a divergence-free (solenoidal) component $\tilde{\mathbf{u}}(k)$ and a gradient (irrotational) component $\tilde{\mathbf{w}}(k)$ parallel to $k$,
\begin{equation}
\tilde{\mathbf{v}}(k)=\tilde{\mathbf{u}}(k)+\tilde{\mathbf{w}}(k), \quad \tilde{\mathbf{w}}(k)=\frac{k(k \cdot \tilde{\mathbf{v}}(k))}{|k|^2} .
\end{equation}
Subtracting the gradient component yields the divergence-free projection,
\begin{equation}
\tilde{\mathbf{u}}(k)=\tilde{\mathbf{v}}(k)-\frac{k(k \cdot \tilde{\mathbf{v}}(k))}{|k|^2}.
\end{equation}

Applied to the Fourier transform of the output of surrogate models $\mathcal{F}(\mathcal{G})$, the divergence-free condition in Fourier space $\mathcal{C}_{div}^*$ is,
\begin{equation}
\mathcal{C}_{\text {div }}^*(\mathcal{F}(\mathcal{G}))=\tilde{\mathbf{u}}=\mathcal{F}(\mathcal{G})-\frac{k(k \cdot \mathcal{F}(\mathcal{G}))}{|k|^2}.
\label{eq16}
\end{equation}

Using the correspondence $\nabla \rightarrow i k$ and $\Delta \rightarrow-|k|^2$, Eq.~\ref{eq16} can be equivalently expressed as,
\begin{equation}
\mathcal{C}_{\text {div }}^*(\mathcal{F}(\mathcal{G}))=\tilde{\mathbf{u}}=\mathcal{F}(\mathcal{G})-\frac{\nabla(\nabla \cdot \mathcal{F}(\mathcal{G}))}{\Delta}.
\end{equation}
\subsection{Mass-conserving projection in the time-dependent systems}
To construct the mass-conserving projection in the time-dependent systems, we consider the output field as a 3D field $\mathbf{u}(\mathbf{x}, \mathbf{t})=\left(\mathbf{u}_\mathbf{t}, \mathbf{u}_{\mathbf{x}_1}, \mathbf{u}_{\mathbf{x}_2}\right)$, where $\mathbf{u}_{\mathbf{t}}$ denotes the temporal flux along $\mathbf{t}$, and $\mathbf{u}_{\mathbf{x}_1}$ and $\mathbf{u}_{\mathbf{x}_2}$ denote the spatial fluxes along $\mathbf{x}_1$ and $\mathbf{x}_2$, respectively. The corresponding divergence-free condition is  expressed as,
\begin{equation}
\frac{\partial \mathbf{u}_{\mathbf{t}}}{\partial \mathbf{t}}+\frac{\partial \mathbf{u}_{\mathbf{x}_1}}{\partial \mathbf{x}_1}+\frac{\partial \mathbf{u}_{\mathbf{x}_2}}{\partial \mathbf{x}_2}=0 .
\end{equation}

Following the derivation presented in Section 7.1, the divergence-free condition in Fourier space $\mathcal{C}_{\text {div }}^*$ can be expressed as,
\begin{equation}
\mathcal{C}_{d i v}^*(\mathcal{F}(\mathcal{G}))=\tilde{\mathbf{u}}=\mathcal{F}(\mathcal{G})-\frac{\tilde{k}(\tilde{k} \cdot \mathcal{F}(\mathcal{G}))}{|\tilde{k}|^2},
\end{equation}
where $\tilde{k}=\left(k_\mathbf{t}, k_1, k_2\right)$ represents the temporal and spatial Fourier modes.

Equivalently, using the Fourier-space correspondence $\nabla=\left(\partial_\mathbf{t}, \partial_{\mathbf{x}_1}, \partial_{\mathbf{x}_2}\right)$ and $\Delta=\partial_\mathbf{t}^2+\partial_{\mathbf{x}_1}^2+\partial_{\mathbf{x}_2}^2$, this can be expressed as,
\begin{equation}
\mathcal{C}_{\text {div }}^*(\mathcal{F}(\mathcal{G}))=\tilde{\mathbf{u}}=\mathcal{F}(\mathcal{G})-\frac{\nabla(\nabla \cdot \mathcal{F}(\mathcal{G}))}{\Delta}.
\end{equation}
\section{Physics-consistent neural operator}
\subsection{Applicability of the physics-consistent projection layer}
The physics-consistent projection layer in PCNO acts as a flexible module, enabling the framework to accommodate a wide range of spatiotemporal dynamical systems. Specifically, for systems governed by momentum conservation (e.g., Kolmogorov flow in vorticity form and flood forecasting), a momentum-conserving projection layer $\mathcal{D}_{\text{mom}}$ is utilized. For systems satisfying both momentum and mass conservation (e.g., Kolmogorov flow in velocity form and 3D atmospheric modeling), the composite projection $\mathcal{D}{\text{mom}} \circ \mathcal{D}{\text{mass}}$ is applied. For systems where neither momentum nor mass is conserved, such as the Kuramoto–Sivashinsky dynamics, the physics-consistent projection layer is omitted.
\subsection{Training strategy of PCNO}
PCNO primarily utilizes a Markov training strategy for long-term spatiotemporal prediction. In this strategy, the model predicts a single future step conditioned on ground-truth solutions from previous steps. For 3D prediction tasks that require outputs directly in both spatial and temporal domains, such as 3D atmospheric modeling, we adopt a one-shot training strategy, where the model takes an entire spatiotemporal grid as input and predicts the corresponding outputs in a single pass.

\section{Diffusion model-enhanced PCNO}
\subsection{DiffPCNO}
\textbf{Consistency training.} 
We train DiffPCNO with the Adam optimizer using learning rate 0.0001. The residuals are normalized to the range [-1, 1] to ensure training stability. The complete algorithm for the consistency training of DiffPCNO is provided in Algorithm~\ref{alg:1}.
\begin{algorithm}
  \caption{Consistency training of DiffPCNO}
  \label{alg:1}
  \KwIn{Dataset $(\mathbf{u},\mathbf{y}) \in \mathcal{D}$, initial  parameters of model $\boldsymbol{f}(\boldsymbol{x}, t ; \boldsymbol{\theta})$, trained PCNO $\mathcal{D} \circ \mathcal{G}$, learning rate, step schedule, $d(., .)$, and $\lambda(\cdot)$.}
  \While{not converge \rm{or} $epoch<max\_epoch$}
  {
  Generate conditional input $\mathbf{\hat{u}}_{t+1}=\mathcal{D} \circ \mathcal{G}\left[\mathbf{u}_t\right]$\;
  Concatenate conditional input $concat(\mathbf{u}_t, \mathbf{\hat{u}}_{t+1})$\;
  Compute normalized residuals  $r_{n}$:
  $r=\mathbf{y}-\mathbf{\hat{u}}_{t+1}$, $r_{n}=(r-r_{\min})/(r_{\max }-r_{\min })$, and $r_{n}=\left(r_{n}-0.5\right) / 0.5$\;
  Determine timesteps $t_i$ and $t_{i+1}$\;
  Sample noise $\mathrm{z} \sim \mathcal{N}(0, \boldsymbol{I})$\;
  $\mathcal{L}_{\mathrm{CT}}\left(\boldsymbol{\theta}, \boldsymbol{\theta}^{-}\right) \longleftarrow$
 $ \mathbb{E}\left[\lambda\left(t_i\right) d\left(\boldsymbol{f}\left(r_{n}+t_{i+1} \mathbf{z}, t_{i+1}, \mathbf{\hat{u}}_{t+1}, \mathbf{u}_t;\boldsymbol{\theta}\right), \boldsymbol{f}\left(r_{n}+t_i \mathbf{z}, t_i, \mathbf{\hat{u}}_{t+1}, \mathbf{u}_t;\boldsymbol{\theta}^{-}\right)\right)\right]$\;
 $epoch = epoch + 1$\; }
    
\end{algorithm}

\textbf{Sampling.} With a well-trained DiffPCNO $\boldsymbol{f}(\boldsymbol{x}, t ; \boldsymbol{\theta})$, samples are generated by initializing with noise $\mathbf{z}$ and computing $\boldsymbol{x}=\boldsymbol{f}\left(\mathbf{z}, t_{max};\boldsymbol{\theta}\right)$. We evaluate DiffPCNO through multiple iterations of alternating denoising and noise injection steps to enhance sample quality. As summarized in Algorithm~\ref{alg:2}, this multistep sampling procedure offers the flexibility to trade computational cost for improved sample fidelity. The time points are selected as $[80.0, 24.4, 5.84, 0.9, 0.661]$.

\begin{algorithm}
  \caption{Sampling of DiffPCNO}
  \label{alg:2}
  \KwIn{Well-trained DiffPCNO $\boldsymbol{f}(\boldsymbol{x}, t ; \boldsymbol{\theta})$, sequence of time points $t_1>t_2>\cdots>t_{N-1}$, initial noise $\hat{\mathbf{x}}_T$.}
  $\mathbf{x} \leftarrow \boldsymbol{f}(\hat{\mathbf{x}}_T, T; \boldsymbol{\theta})$ \;
  \For{$n=1 \text { to } N-1$}
      {   
          Sample $\mathbf{z} \sim \mathcal{N}(0, \boldsymbol{I})$ \;
          $\hat{\mathbf{x}}_{t_n} \leftarrow \mathbf{x}+\sqrt{t_n^2-0.002^2} \mathbf{z}$ \;
          $\mathbf{x} \leftarrow \boldsymbol{f}(\hat{\mathbf{x}}_{t_n}, t_n; \boldsymbol{\theta})$.         
      }  
   Obtain residuals $r_{n}=\mathbf{x}$ and clamp to [0,1] \;
   Denormalize residuals: $r_{n}=\left(r_{n}\right)*0.5+0.5$, then $r=r_{n}*(r_{\max }-r_{\min })+r_{\min}$
\end{algorithm}

\subsection{PCNO-Refiner}
\textbf{Consistency training.} 
As shown in Supplementary Fig.~\ref{fig:s2}, PCNO-Refiner conditions on the deterministic prediction $\mathbf{\hat{u}}_{t+1}$ from PCNO and the current state $\mathbf{u}_t$,  aiming to produce predictions closer to the ground truth $\mathbf{y}$, using the consistency training. We train the PCNO-Refiner using the Adam optimizer with a learning rate of 0.0001. The training data are normalized to the range [-1, 1]. The complete procedure for the consistency training of PCNO-Refiner is detailed in Algorithm~\ref{alg:3}.
\begin{algorithm}
  \caption{Consistency training of PCNO-Refiner}
  \label{alg:3}
  \KwIn{Dataset $(\mathbf{u},\mathbf{y}) \in \mathcal{D}$, initial  parameters of model $\boldsymbol{f}(\boldsymbol{x}, t ; \boldsymbol{\theta})$, trained PCNO $\mathcal{D} \circ \mathcal{G}$, learning rate, step schedule, $d(., .)$, and $\lambda(\cdot)$.}
  \While{not converge \rm{or} $epoch<max\_epoch$}
  {
  Generate conditional input $\mathbf{\hat{u}}_{t+1}=\mathcal{D} \circ \mathcal{G}\left[\mathbf{u}_t\right]$\;
  Concatenate conditional input $concat(\mathbf{u}_t, \mathbf{\hat{u}}_{t+1})$\;
  Determine timesteps $t_i$ and $t_{i+1}$\;
  Sample noise $\mathrm{z} \sim \mathcal{N}(0, \boldsymbol{I})$\;
  $\mathcal{L}_{\mathrm{CT}}\left(\boldsymbol{\theta}, \boldsymbol{\theta}^{-}\right) \longleftarrow$
 $ \mathbb{E}\left[\lambda\left(t_i\right) d\left(\boldsymbol{f}\left(\mathbf{y}+t_{i+1} \mathbf{z}, t_{i+1}, \mathbf{\hat{u}}_{t+1}, \mathbf{u}_t;\boldsymbol{\theta}\right), \boldsymbol{f}\left(\mathbf{y}+t_i \mathbf{z}, t_i, \mathbf{\hat{u}}_{t+1}, \mathbf{u}_t;\boldsymbol{\theta}^{-}\right)\right)\right]$\;
 $epoch = epoch + 1$\; }
    
\end{algorithm}

\textbf{Sampling.} The sampling process of PCNO-Refiner closely follows that of DiffPCNO, as both employ a multistep sampling strategy. However, unlike DiffPCNO, PCNO-Refiner generates the refined predictions at each iteration, as summarized in Algorithm~\ref{alg:4}. The sampling is performed at the time points 
$[80.0, 24.4, 5.84, 0.9, 0.661]$.

\begin{algorithm}
  \caption{Sampling of PCNO-Refiner}
  \label{alg:4}
  \KwIn{Well-trained DiffPCNO $\boldsymbol{f}(\boldsymbol{x}, t ; \boldsymbol{\theta})$, sequence of time points $t_1>t_2>\cdots>t_{N-1}$, initial noise $\hat{\mathbf{x}}_T$.}
  $\mathbf{x} \leftarrow \boldsymbol{f}(\hat{\mathbf{x}}_T, T; \boldsymbol{\theta})$ \;
  \For{$n=1 \text { to } N-1$}
      {   
          Sample $\mathbf{z} \sim \mathcal{N}(0, \boldsymbol{I})$ \;
          $\hat{\mathbf{x}}_{t_n} \leftarrow \mathbf{x}+\sqrt{t_n^2-0.002^2} \mathbf{z}$ \;
          $\mathbf{x} \leftarrow \boldsymbol{f}(\hat{\mathbf{x}}_{t_n}, t_n; \boldsymbol{\theta})$.         
      }  
   Generate refined predictions $\mathbf{u}_{t+1}=\mathbf{x}$ and denormalize
\end{algorithm}

\subsection{Model architecture}
For the model $\boldsymbol{F}(\boldsymbol{x}, t; \boldsymbol{\theta})$, we employ a modern U-Net architecture~\cite{gupta2023towards}. In DiffPCNO, the deterministic prediction $\mathbf{\hat{u}}_{t+1}$ from PCNO and the current state $\mathbf{u}_t$ are concatenated with $(\mathbf{y} - \mathbf{\hat{u}}_{t+1}) + t \mathbf{z}$ as the model input. PCNO-Refiner takes $\mathbf{\hat{u}}_{t+1}$, $\mathbf{u}_t$, and $\mathbf{y} + t \mathbf{z}$ as input channels.

The U-Net architecture comprises an encoder and a decoder, each implemented with multiple pre-activation ResNet blocks connected through skip connections between corresponding encoder and decoder levels. Each ResNet block consists of Group Normalization, GELU activation functions, and convolutional layers with a kernel size of 3. Conditioning parameters are projected into a feature vector space using sinusoidal embeddings, as commonly applied in Transformer architectures~\cite{vaswani2017attention}. These conditioning variables include the discrete time step $t$ in consistency training across all experiments, as well as $\Delta t$ and $\Delta x$ for the 1D KSE with fixed viscosity, and additionally the viscosity coefficient $\nu$ for the 1D KSE with varying viscosity. We integrate the feature vectors obtained from conditioning parameters via linear layers into the U-Net through AdaGN layers~\cite{nichol2021improved,lippe2023pde}, which predict channel-wise scale and shift parameters applied after the  Group Normalization operation in each residual block. Furthermore, we employ attention mechanisms within the residual blocks, similar to multi-head self-attention in Transformer models~\cite{vaswani2017attention}. In addition, we employ a 1D U-Net for the 1D KSE and a 2D U-Net for 2D Kolmogorov turbulent flow, 2D flood forecasting, and 2D atmospheric modeling.

\newpage
\section*{Supplementary Figures}
\setcounter{figure}{0}
\setcounter{table}{0}
\renewcommand{\thefigure}{\arabic{figure}}
\renewcommand{\thetable}{\arabic{table}}
\captionsetup[figure]{name=Supplementary Fig}
\captionsetup[table]{name=Supplementary Table}

\begin{figure}[htp!]
	\centering
	{\includegraphics[width = 1.0\textwidth]{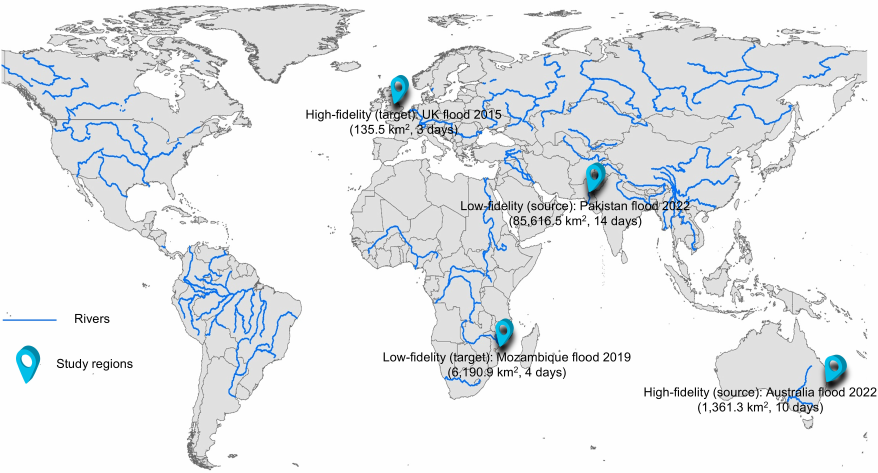}}
	\caption{Locations of the study areas.
	}
	\label{fig:s1}
\end{figure}

\begin{figure}[htp!]
	\centering
	{\includegraphics[width = 1.0\textwidth]{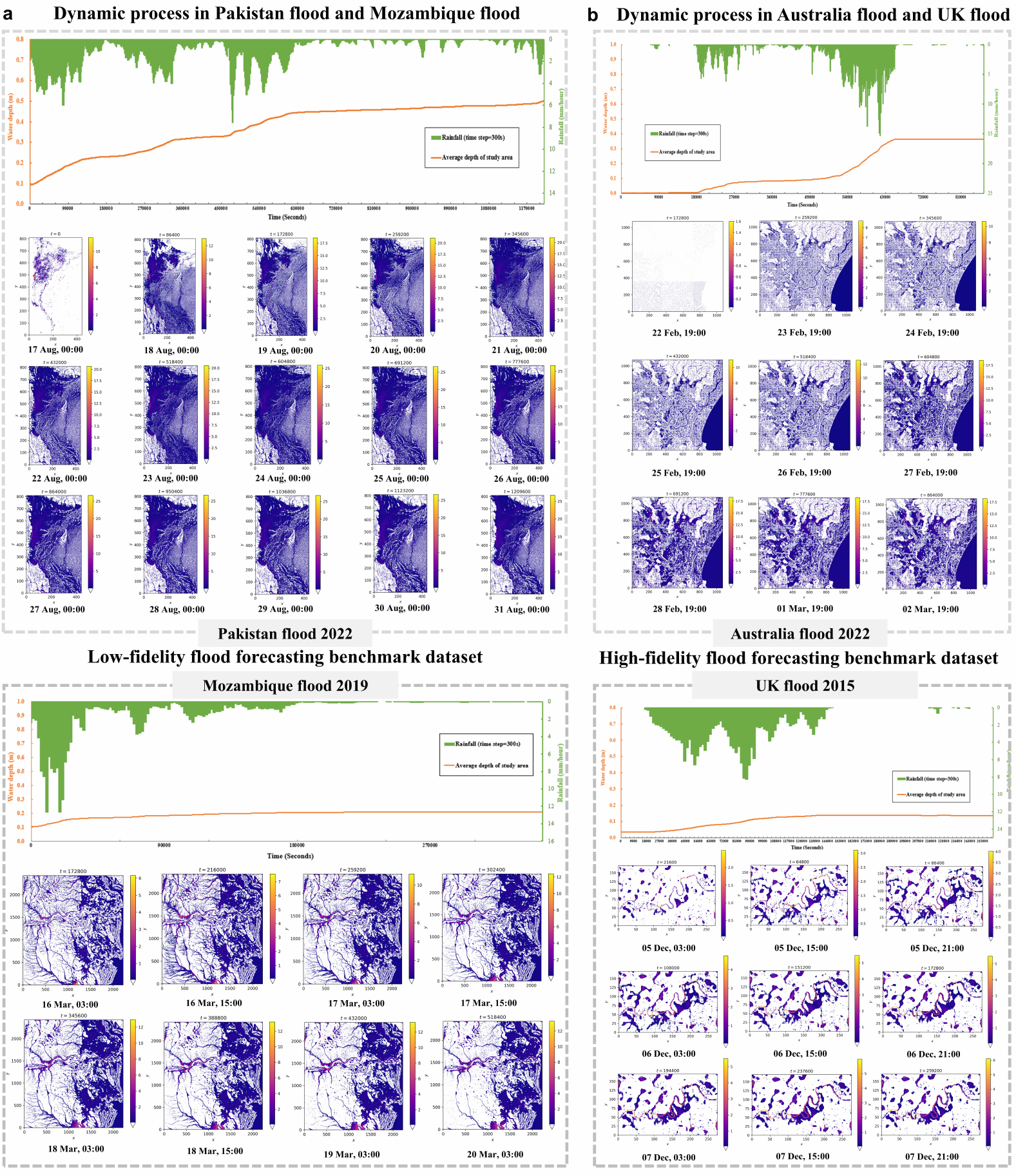}}
	\caption{FloodCastBench dataset for both low-fidelity and high-fidelity flood forecasting. \textbf{a,} Benchmark dataset for low-fidelity flood forecasting. \textbf{b,} Benchmark dataset for high-fidelity flood forecasting.
	}
	\label{fig:s02}
\end{figure}

\begin{figure}[htp!]
	\centering
	{\includegraphics[width = 1.1\textwidth]{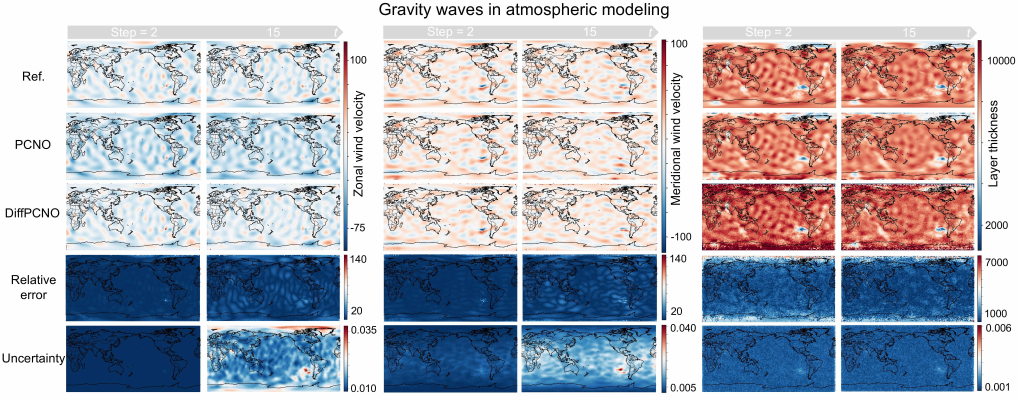}}
	\caption{Visualization and uncertainty quantification using 2D models for learning  the transformed variables $(u_x h, u_y h \sin \theta, R h \sin \theta)$ (15 min temporal, T63 spectral truncation, 15 time steps). The atmospheric variables $(u_x, u_y, h)$ are recovered from the outputs $(u_x h, u_y h \sin \theta, R h \sin \theta)$.
	}
	\label{fig:s0}
\end{figure}

\begin{figure}[htp!]
	\centering
	{\includegraphics[width = 1.0\textwidth]{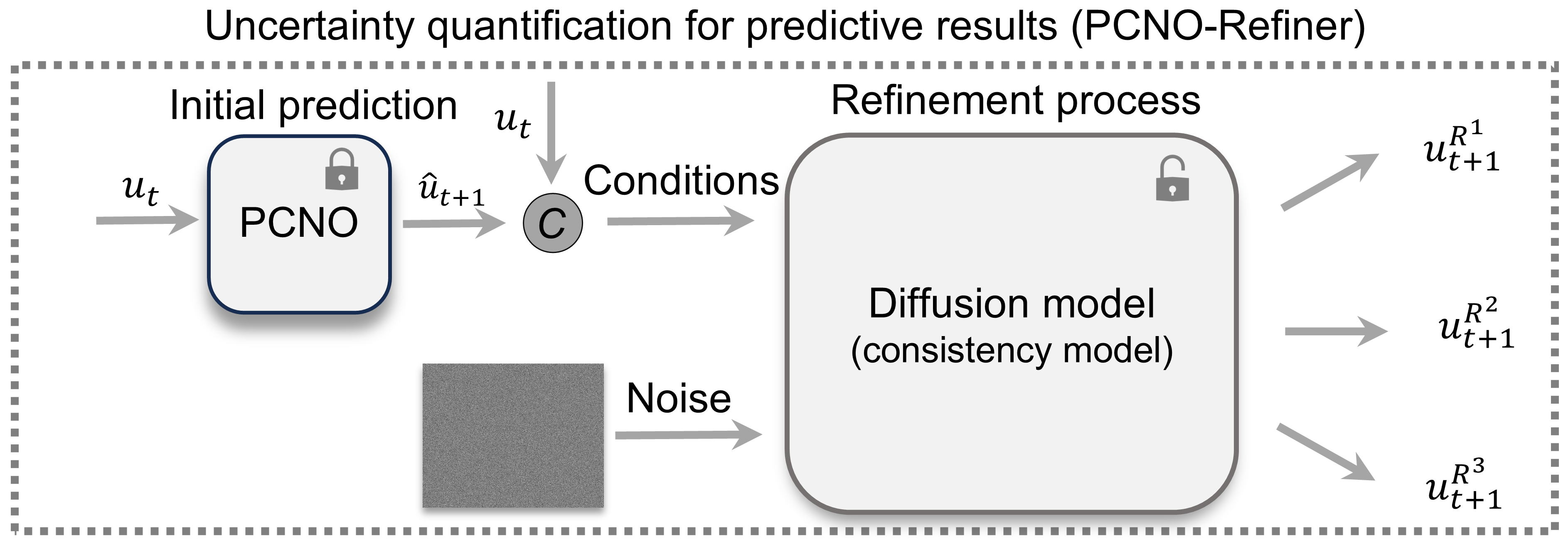}}
	\caption{Framework of PCNO-Refiner.
	}
	\label{fig:s2}
\end{figure}

\clearpage
\newpage
\section*{Supplementary Tables}
\begin{table}[htbp!]
\centering
\caption{Complete list of symbols in the Methods}
\label{table1}
\begin{tabular}{p{0.3\textwidth} p{0.7\textwidth}}
\hline
\textbf{Symbols} & \multicolumn{1}{c}{\textbf{Meaning}} \\
\hline
\multicolumn{2}{c}{Spatiotemporal dynamics processes} \\
$\Omega \subset \mathbb{R}^d$ & $d$-dimensional spatial domain. \\
$\mathcal{T} \subset \mathbb{R}$ & Temporal domain. \\
$(\mathbf{x}, \mathbf{t})$ & Points in the spatiotemporal domain, $(\mathbf{x}, \mathbf{t}) \in \Omega \times \mathcal{T}$. \\
$\mathbf{u}(\mathbf{x}, \mathbf{t})$ & State variable $\mathbf{u}(\mathbf{x}, \mathbf{t}) \in  \mathbb{R}^m$ with $m$ components defined over the spatiotemporal domain. \\
$\nabla_{\mathbf{x}}$ & Nabla operator with respect to the spatial coordinate $\mathbf{x}$. \\
$F(\cdot)$ & Nonlinear function describing the right-hand side of PDEs. \\
$\mathcal{I}(\mathbf{u}; t=0, \mathbf{x} \in \Omega)=0$ & Initial condition. \\
$\mathcal{B}\left(\mathbf{u}, \nabla_{\mathbf{x}} \mathbf{u}, \cdots ; \mathbf{x} \in \partial \Omega\right)=0$ & Boundary condition. $\partial \Omega$ represents the boundary of the system domain. \\
\hline
\multicolumn{2}{c}{Surrogate model $\mathcal{G}$} \\
$\theta$ & Parameters of the NO surrogate model. \\
 \\
 $\mathcal{P}$, $\mathcal{Q}$ & Pixel-wise transformations using multilayer perceptron. \\
$\left\{\mathcal{J}_1, \mathcal{J}_2, \ldots, \mathcal{J}_L\right\}$ & $L$-layer nonlinear operators. \\
$W_l$ & A matrix functioning as a linear transformation. \\
$\mathcal{K}_l$ & An integral kernel operator. \\
$\sigma$ & Nonlinear activation function. \\
$v_l, v_{l+1}$ & The feature functions at the $l$ and $l+1$th layers of the nonlinear operator. \\
$\mathcal{F}$ & Fast Fourier Transform (FFT). \\
$\mathcal{F}^{-1}$ & The inverse FFT. \\
$R_\phi$ & A parametric function to realize the Fourier transform of a periodic function. \\
$\phi$ & The learnable parameters of $R_\phi$. \\
$k$ & Fourier modes. \\
\hline
\multicolumn{2}{c}{Physics-consistent projection layer $\mathcal{D}$} \\
$\mathcal{C}$ & A set of physical constraints (such as mass and momentum conservations).\\
$\mathcal{D}^*$ & The imposition of physical constraints in Fourier space.\\
$L_g[\mathcal{G}]$ & Translation on the
output of the surrogate model $\mathcal{G}$, defined as $L_g[\mathcal{G}] =\tilde{\mathcal{G}}$ with $\tilde{\mathcal{G}}(x+g)=\mathcal{G}(x)$.\\
$g \in \mathbb{R}^d$ & Translation vector defined in the $d$-dimensional output space.\\
$L_R[\mathcal{G}]$ & Rotation on the
output of the surrogate model $\mathcal{G}$, defined as $L_R[\mathcal{G}] = \tilde{\mathcal{G}}$ with $\tilde{\mathcal{G}}(R\mathbf{x})=R\mathcal{G}(\mathbf{x})$. \\
$R \in \mathbb{R}^{d_m \times d_m}$ &  Rotation transformation matrix. $R$ is an orthogonal matrix, where $d_m$ denotes the dimension of the matrix.\\
$W$ & Kernel filters.\\
$W_{inv}$ & Invariant convolutions with rotational symmetry. \\
$L_R[\mathcal{F}(W)]$ & Rotation-invariant kernel.\\
$\rho$ & Fluid density.\\
$\mu$ & Mass flux. $\mu=\rho \mathbf{u}$, where  $\mathbf{u}$ is the velocity field.\\
$\tilde{\mathbf{u}}$ & Divergence-free field in Fourier space.\\
$\mathcal{C}_{div}^*$ & Divergence-free condition in Fourier space.\\
$\nabla(\nabla \cdot \mathcal{F}(\mathcal{G}))$ & Gradient of the divergence of $\mathcal{F}(\mathcal{G})$.\\
$\Delta$ & Laplacian in the frequency domain.\\
$W_{spe}$ & Spectral convolution.\\
$\mathcal{D}_{mass}$ & Mass-conserving projection.\\
\hline
\end{tabular}
\end{table}
\begin{table}[ht!]
\centering
\begin{tabular}{p{0.3\textwidth} p{0.7\textwidth}}
\hline
\multicolumn{2}{c}{Embedding uncertainty to learn spatiotemporal dynamics processes} \\
$\mathcal{P}\left[\theta_g\right]$ & Generative consistency model parameterized by $\theta_g$.\\
$\mathbf{r}$ & Residual error between predictive and ground-truth solutions. \\
\hline
\multicolumn{2}{c}{Consistency models} \\
$\boldsymbol{f}(\boldsymbol{x}, t)$ & Consistency function.\\
$t_{min}, t_{max}$ & Time interval.\\
$\boldsymbol{f}\left(\boldsymbol{x}_\epsilon, \epsilon\right)=\boldsymbol{x}_\epsilon$ & Boundary conditions of the consistency function.\\
$\boldsymbol{f}(\boldsymbol{x}, t;\boldsymbol{\theta})$ & Neural network trained to approximate the target consistency function.\\
$\boldsymbol{F}(\boldsymbol{x}, t;\boldsymbol{\theta})$ & U-Net.\\
$c_{\text {skip }}(t), c_{\text {out }}(t)$ & Differentiable functions such that $c_{\text {skip }}\left(t_{\min }\right)=1$ and $c_{\text {out }}\left(t_{\min }\right)=0$\\
$t_{\min }=t_1,t_2,\cdots,t_N=t_{\max}$ & A sequence of time steps in consistency training.\\
$\operatorname{erf}$ & Error function.\\
$p(i)$ & Discretized lognormal distribution.\\
$\mathrm{k}$ & Current training step for an improved discretization curriculum.\\
$\mathrm{K}$ & The total number of steps for an improved discretization curriculum.\\
$s_0, s_1$ & The initial and maximum discretization steps.\\
$\mathcal{L}_{\mathrm{CT}}$ & Consistency training
loss.\\
$\boldsymbol{\theta}, \boldsymbol{\theta}^{-}$ & The student and teacher network weights.\\
$\lambda\left(t_i\right)$ & Weighting function.\\
$d(x, y)$ & Pseudo-Huber metric.\\
$c$ & Adjustable constant.\\
$\mathbf{z}$ & Noise.\\
\hline
\multicolumn{2}{c}{Diffusion model-enhanced PCNO} \\
$\mathbf{\hat{u}}$ & Deterministic prediction from PCNO.\\
$\mathbf{y}$ & Ground-truth solution.\\
\hline
\multicolumn{2}{c}{Evaluation metrics} \\
$\hat{\mathbf{y}}$ & Predicted solution.\\
$n$ & The number of test samples.\\
$\|\cdot\|_2$ & $L_2$ norm.\\
$r$ & Pearson correlation coefficient between predicted and true values.\\
$\bar{\mathbf{y}}$ & Mean values of the true solutions.\\
$\hat{\bar{\mathbf{y}}}$ & Mean values of the predicted solutions.\\
$\mathcal{L}_{\mathrm{div}}$ & Divergence loss.\\
$N$ & The total number of spatial points.\\
$|\cdot|$ & The absolute value. \\
$\nabla \cdot \mathbf{u}_{\mathrm{pred}, i}$ & Divergence of the predicted vector field at the $i$-th spatial point. \\
$\mathcal{L}_{\mathrm{M}}$ & Momentum loss. \\
$M_{\mathrm{pred}}$ & Predicted momentum. \\
$M_{\mathrm{ref}}$ & Reference momentum. \\
$\gamma$ & Threshold in CSI. \\
$\mathrm{TP}$ & True positives (cells where both predictions and ground truths exceed $\gamma$ ). \\
$\mathrm{FP}$ & False positives (cells where ground truths are below $\gamma$ but predictions exceed $\gamma$ ). \\
$\mathrm{FN}$ & False negatives (cells where the model fails to predict a flooded area). \\
\hline
\end{tabular}
\end{table}

\begin{table}[htbp]
\centering
\caption{Training settings for 1D KSE.}
\label{table0}
\begin{tabular}{ccc}
\hline
\textbf{Methods}                             & \textbf{Parameters}                & \textbf{Values}  \\
\hline
\multirow{9}{*}{\textbf{U-Net}}              & Data   normalization               & {[}0,1{]}        \\
                                             & Epochs                             & 100              \\
                                             & Batch size                         & 128              \\
                                             & Optimizer                          & Adam             \\
                                             & Learning rate                      & 0.001            \\
                                             & Scheduler                          & Cosine Annealing \\
                                             & Weight decay                       & 0.0001           \\
                                             & Network                            & Modern U-Net     \\
                                             & Hidden channels                    & 64               \\
                                             \hline
\multirow{10}{*}{\textbf{PCNO (FNO)}}        & Data   normalization               & {[}0,1{]}        \\
                                             & Epochs                             & 100              \\
                                             & Batch size                         & 128              \\
                                             & Optimizer                          & Adam             \\
                                             & Learning rate                      & 0.001            \\
                                             & Scheduler                          & Cosine Annealing \\
                                             & Weight decay                       & 0.0001           \\
                                             & Fourier layers                     & 4                \\
                                             & Fourier modes                      & 12               \\
                                             & Width                              & 20               \\
                                             \hline
\multirow{13}{*}{\textbf{PDE-Refiner}}       & Data   normalization               & {[}0,1{]}        \\
                                             & Epochs                             & 400              \\
                                             & Batch size                         & 128              \\
                                             & Optimizer                          & Adam             \\
                                             & Learning rate                      & 0.001            \\
                                             & Scheduler                          & Cosine Annealing \\
                                             & Weight decay                       & 0.0001           \\
                                             & Network                            & FNO              \\
                                             & Predict difference                 & TRUE             \\
                                             & Difference weight                  & 0.3              \\
                                             & Minimum noise standard   deviation & 4.00E-07         \\
                                             & EMA Decay                          & 0.995            \\
                                             & Number of refinement steps         & 3                \\
                                             \hline
\end{tabular}%
\end{table}
\begin{table}[ht!]
\centering
\begin{tabular}{cccc}
\hline
\multirow{13}{*}{\textbf{PDE-Refiner+}}      & Data   normalization               & {[}0,1{]}        \\
                                             & Epochs                             & 400              \\
                                             & Batch size                         & 128              \\
                                             & Optimizer                          & Adam             \\
                                             & Learning rate                      & 0.001            \\
                                             & Scheduler                          & Cosine Annealing \\
                                             & Weight decay                       & 0.0001           \\
                                             & Network                            & FNO              \\
                                             & Predict difference                 & TRUE             \\
                                             & Difference weight                  & 0.3              \\
                                             & Minimum noise standard   deviation & 4.00E-07         \\
                                             & EMA Decay                          & 0.995            \\
                                             & Number of refinement steps         & 3                \\
                                             \hline
\multirow{10}{*}{\textbf{Consistency model}} & Data   normalization               & {[}-1,1{]}       \\
                                             & Epochs                             & 100              \\
                                             & Batch size                         & 128              \\
                                             & Optimizer                          & Adam             \\
                                             & Learning rate                      & 0.001            \\
                                             & Scheduler                          & LinearLR         \\
                                             & Betas                              & (0.9, 0.995)     \\
                                             & Start\_factor                      & 1.00E-05         \\
                                             & Total\_iters                       & 1000             \\
                                             & Network                            & Modern U-Net     \\
                                             \hline
\multirow{10}{*}{\textbf{PCNO-Refiner}}      & Data   normalization               & {[}-1,1{]}       \\
                                             & Epochs                             & 400              \\
                                             & Batch size                         & 128              \\
                                             & Optimizer                          & Adam             \\
                                             & Learning rate                      & 0.0001           \\
                                             & Scheduler                          & LinearLR         \\
                                             & Betas                              & (0.9, 0.995)     \\
                                             & Start\_factor                      & 1.00E-05         \\
                                             & Total\_iters                       & 1000             \\
                                             & Network                            & Modern U-Net     \\
                                             \hline
\multirow{10}{*}{\textbf{DiffPCNO}}          & Data   normalization               & {[}0,1{]}        \\
                                             & Epochs                             & 100              \\
                                             & Batch size                         & 128              \\
                                             & Optimizer                          & Adam             \\
                                             & Learning rate                      & 0.0001           \\
                                             & Scheduler                          & LinearLR         \\
                                             & Betas                              & (0.9, 0.995)     \\
                                             & Start\_factor                      & 1.00E-05         \\
                                             & Total\_iters                       & 1000             \\
                                             & Network                            & Modern U-Net   \\
                                             \hline
\end{tabular}
\end{table}

\begin{table}[htbp]
\centering
\caption{Training settings for Kolmogorov turbulent flow.}
\label{table02}
\begin{tabular}{cccc}
\hline
\textbf{Methods}                             & \textbf{Parameters}                    & \textbf{Velocity form} & \textbf{Vorticity form} \\
\hline
\multirow{10}{*}{\textbf{LNO}}               & Epochs                                 & 100                                & 100                                 \\
                                             & Batch size                             & 20                                 & 20                                  \\
                                             & Optimizer                              & Adam                               & Adam                                \\
                                             & Learning rate                          & 0.001                              & 0.001                               \\
                                             & Scheduler                              & Cosine Annealing                   & Cosine Annealing                    \\
                                             & Weight decay                           & 0.0001                             & 0.0001                              \\
                                             & Width                                  & 20                                 & 20                                  \\
                                             & Modes                                  & 12                                 & 12                                  \\
                                             & Laplace layer                          & 1                                  & 1                                   \\
                                             & Number of output channels              & 2                                  & 1                                   \\
                                             \hline
\multirow{8}{*}{\textbf{U-Net}}              & Epochs                                 & 100                                & 100                                 \\
                                             & Batch size                             & 20                                 & 20                                  \\
                                             & Optimizer                              & Adam                               & Adam                                \\
                                             & Learning rate                          & 0.001                              & 0.001                               \\
                                             & Learning rate scheduler                & Cosine Annealing                   & Cosine Annealing                    \\
                                             & Weight decay                           & 0.0001                             & 0.0001                              \\
                                             & Feature dimension of the first   layer & 32                                 & 32                                  \\
                                             & Number of output channels              & 2                                  & 1                                   \\
                                             \hline
\multirow{10}{*}{\textbf{ClawFNO}}           & Epochs                                 & 100                                & -                                   \\
                                             & Batch size                             & 20                                 & -                                   \\
                                             & Optimizer                              & Adam                               & -                                   \\
                                             & Learning rate                          & 0.001                              & -                                   \\
                                             & Scheduler                              & Cosine Annealing                   & -                                   \\
                                             & Weight decay                           & 0.0001                             & -                                   \\
                                             & Fourier layers                         & 4                                  & -                                   \\
                                             & Fourier modes                          & 12                                 & -                                   \\
                                             & Width                                  & 20                                 & -                                   \\
                                             & Number of output channels              & 1                                  & -                                   \\
                                             \hline
\multirow{10}{*}{\textbf{G-FNO}}             & Epochs                                 & 100                                & 100                                 \\
                                             & Batch size                             & 20                                 & 20                                  \\
                                             & Optimizer                              & Adam                               & Adam                                \\
                                             & Learning rate                          & 0.001                              & 0.001                               \\
                                             & Learning rate scheduler                & Cosine Annealing                   & Cosine Annealing                    \\
                                             & Weight decay                           & 0.0001                             & 0.0001                              \\
                                             & Fourier layers                         & 4                                  & 4                                   \\
                                             & Fourier modes                          & 12                                 & 12                                  \\
                                             & Width                                  & 10                                 & 10                                  \\
                                             & Number of output channels              & 2                                  & 1                                   \\
                                             \hline

\multirow{10}{*}{\textbf{FNO}}               & Epochs                                 & 100                                & 100                                 \\
                                             & Batch size                             & 20                                 & 20                                  \\
                                             & Optimizer                              & Adam                               & Adam                                \\
                                             & Learning rate                          & 0.001                              & 0.001                               \\
                                             & Scheduler                              & Cosine Annealing                   & Cosine Annealing                    \\
                                             & Weight decay                           & 0.0001                             & 0.0001                              \\
                                             & Fourier layers                         & 4                                  & 4                                   \\
                                             & Fourier modes                          & 12                                 & 12                                  \\
                                             & Width                                  & 20                                 & 20                                  \\
                                             & Number of output channels              & 2                                  & 1                                   \\
                                             \hline
                                             \end{tabular}%
\end{table}
\begin{table}[ht!]
\centering
\begin{tabular}{cccc}
\hline
\multirow{10}{*}{\textbf{PCNO w/o Momentum}} & Epochs                                 & 100                                & -                                   \\
                                             & Batch size                             & 20                                 & -                                   \\
                                             & Optimizer                              & Adam                               & -                                   \\
                                             & Learning rate                          & 0.001                              & -                                   \\
                                             & Scheduler                              & Cosine Annealing                   & -                                   \\
                                             & Weight decay                           & 0.0001                             & -                                   \\
                                             & Fourier layers                         & 4                                  & -                                   \\
                                             & Fourier modes                          & 12                                 & -                                   \\
                                             & Width                                  & 20                                 & -                                   \\
                                             & Number of output channels              & 2                                  & -                                   \\
                                             \hline
\multirow{10}{*}{\textbf{PCNO w/o Mass}}     & Epochs                                 & 100                                & -                                   \\
                                             & Batch size                             & 20                                 & -                                   \\
                                             & Optimizer                              & Adam                               & -                                   \\
                                             & Learning rate                          & 0.001                              & -                                   \\
                                             & Scheduler                              & Cosine Annealing                   & -                                   \\
                                             & Weight decay                           & 0.0001                             & -                                   \\
                                             & Fourier layers                         & 4                                  & -                                   \\
                                             & Fourier modes                          & 12                                 & -                                   \\
                                             & Width                                  & 20                                 & -                                   \\
                                             & Number of output channels              & 2                                  & -                                   \\
                                             \hline
\multirow{10}{*}{\textbf{PCNO}}              & Epochs                                 & 100                                & 100                                 \\
                                             & Batch size                             & 20                                 & 20                                  \\
                                             & Optimizer                              & Adam                               & Adam                                \\
                                             & Learning rate                          & 0.001                              & 0.001                               \\
                                             & Scheduler                              & Cosine Annealing                   & Cosine Annealing                    \\
                                             & Weight decay                           & 0.0001                             & 0.0001                              \\
                                             & Fourier layers                         & 4                                  & 4                                   \\
                                             & Fourier modes                          & 12                                 & 12                                  \\
                                             & Width                                  & 20                                 & 20                                  \\
                                             & Number of output channels              & 2                                  & 1                                   \\
                                             \hline
\multirow{12}{*}{\textbf{Consistency model}} & Data   normalization                   & {[}-1,1{]}                         & {[}-1,1{]}                          \\
                                             & Epochs                                 & 100                                & 100                                 \\
                                             & Batch size                             & 20                                 & 20                                  \\
                                             & Optimizer                              & Adam                               & Adam                                \\
                                             & Learning rate                          & 0.0001                             & 0.0001                              \\
                                             & Scheduler                              & LinearLR                           & LinearLR                            \\
                                             & Betas                                  & (0.9, 0.995)                       & (0.9, 0.995)                        \\
                                             & Start\_factor                          & 1.00E-05                           & 1.00E-05                            \\
                                             & Total\_iters                           & 1000                               & 1000                                \\
                                             & Network                                & Modern U-Net                       & Modern U-Net                        \\
                                             & Hidden channels                        & 64                                 & 64                                  \\
                                             & Number of output channels              & 2                                  & 1                                   \\
                                             \hline
\multirow{12}{*}{\textbf{DiffPCNO}}          & Data   normalization                   & {[}0,1{]}                          & {[}0,1{]}                           \\
                                             & Epochs                                 & 100                                & 100                                 \\
                                             & Batch size                             & 20                                 & 20                                  \\
                                             & Optimizer                              & Adam                               & Adam                                \\
                                             & Learning rate                          & 0.0001                             & 0.0001                              \\
                                             & Scheduler                              & LinearLR                           & LinearLR                            \\
                                             & Betas                                  & (0.9, 0.995)                       & (0.9, 0.995)                        \\
                                             & Start\_factor                          & 1.00E-05                           & 1.00E-05                            \\
                                             & Total\_iters                           & 1000                               & 1000                                \\
                                             & Network                                & Modern U-Net                       & Modern U-Net                        \\
                                             & Hidden channels                        & 64                                 & 64                                  \\
                                             & Number of output channels              & 2                                  & 1  \\
                                             \hline
\end{tabular}
\end{table}

\begin{table}[h!]
\centering
\caption{Training settings for real-world flood inundation forecasting.}
\label{table04}
\begin{tabular}{cccc}
\hline
\textbf{Methods}                                                                                            & \textbf{Parameters}       & \textbf{Pakistan flood 2022}                                 & \textbf{Australia flood 2022}                                \\ \hline
\multirow{11}{*}{\textbf{\makecell{FNO without rainfall \\ and terrain DEM; \\  PCNO without rainfall   \\and terrain DEM}}} & Epochs                    & 100                                                          & 100                                                          \\
                                                                                                            & Batch size                & 2                                                            & 1                                                            \\
                                                                                                            & Optimizer                 & Adam                                                         & Adam                                                         \\
                                                                                                            & Learning rate             & 0.001                                                        & 0.001                                                        \\
                                                                                                            & Scheduler                 & Cosine Annealing                                             & Cosine Annealing                                             \\
                                                                                                            & Weight decay              & 0.0001                                                       & 0.0001                                                       \\
                                                                                                            & Fourier layers            & 4                                                            & 4                                                            \\
                                                                                                            & Fourier modes             & 12                                                           & 12                                                           \\
                                                                                                            & Width                     & 20                                                           & 20                                                           \\
                                                                                                            & Number of input channels  & 3                                                            & 3                                                            \\
                                                                                                            & Number of output channels & 1                                                            & 1                                                            \\ \hline
\multirow{11}{*}{\textbf{FNO; PCNO}}                                                                     & Epochs                    & 100                                                          & 100                                                          \\
                                                                                                            & Batch size                & 2                                                            & 1                                                            \\
                                                                                                            & Optimizer                 & Adam                                                         & Adam                                                         \\
                                                                                                            & Learning rate             & 0.001                                                        & 0.001                                                        \\
                                                                                                            & Scheduler                 & Cosine Annealing                                             & Cosine Annealing                                             \\
                                                                                                            & Weight decay              & 0.0001                                                       & 0.0001                                                       \\
                                                                                                            & Fourier layers            & 4                                                            & 4                                                            \\
                                                                                                            & Fourier modes             & 12                                                           & 12                                                           \\
                                                                                                            & Width                     & 20                                                           & 20                                                           \\
                                                                                                            & Number of input channels  & 5                                                            & 5                                                            \\
                                                                                                            & Number of output channels & 1                                                            & 1                                                            \\ \hline
\multirow{13}{*}{\textbf{DiffPCNO}}                                                                         & Data normalization        & \makecell{Scale water   depths to {[}0,1{]} \\ using log1p and global max} & \makecell{Scale water   depths to {[}0,1{]} using \\ log1p and global max} \\
                                                                                                            & Epochs                    & 100                                                          & 100                                                          \\
                                                                                                            & Batch size                & 2                                                            & 1                                                            \\
                                                                                                            & Optimizer                 & Adam                                                         & Adam                                                         \\
                                                                                                            & Learning rate             & 0.0001                                                       & 0.0001                                                       \\
                                                                                                            & Scheduler                 & LinearLR                                                     & LinearLR                                                     \\
                                                                                                            & Betas                     & (0.9, 0.995)                                                 & (0.9, 0.995)                                                 \\
                                                                                                            & Start\_factor             & 1.00E-05                                                     & 1.00E-05                                                     \\
                                                                                                            & Total\_iters              & 1000                                                         & 1000                                                         \\
                                                                                                            & Network                   & Modern U-Net                                                 & Modern U-Net                                                 \\
                                                                                                            & Hidden channels           & 64                                                           & 64                                                           \\
                                                                                                            & Number of input channels  & 7                                                            & 7                                                            \\
                                                                                                            & Number of output channels & 1                                                            & 1                                                            \\ \hline
\end{tabular}
\end{table}

\begin{table}[htbp]
\centering
\caption{Training settings for atmospheric modeling.}
\label{table2}
\begin{tabular}{cccc}
\hline
\textbf{Methods} & \textbf{Parameters} & \textbf{Transformed variables} & \textbf{Atmospheric variables} \\
\hline
\multirow{9}{*}{\textbf{U-Net-2D}}   & Data normalization & [0,1] & [0,1] \\
& Training strategies & Markov & Markov \\
& Epochs & 100 & 100 \\
& Batch size & 10 & 10 \\
& Optimizer & Adam & Adam \\
& Learning rate & 0.001 & 0.001 \\
& Learning rate scheduler & Cosine Annealing & Cosine Annealing \\
& Weight decay & 0.0001 & 0.0001 \\
& Feature dimension of the first layer & 32 & 32 \\
\hline
\multirow{11}{*}{\textbf{FNO-2D}} & Data normalization & [0,1] & [0,1] \\
& Training strategies & Markov & Markov \\
& Epochs & 100 & 100 \\
& Batch size & 10 & 10 \\
& Optimizer & Adam & Adam \\
& Learning rate & 0.001 & 0.001 \\
& Learning rate scheduler & Cosine Annealing & Cosine Annealing \\
& Weight decay & 0.0001 & 0.0001 \\
& Fourier layers & 4 & 4 \\
& Spatial Fourier modes & 22 & 22 \\
& Width & 20 & 20 \\
\hline
\multirow{11}{*}{\textbf{G-FNO-2D}} & Data normalization & [0,1] & [0,1] \\
& Training strategies & Markov & Markov \\
& Epochs & 100 & 100 \\
& Batch size & 10 & 10 \\
& Optimizer & Adam & Adam \\
& Learning rate & 0.001 & 0.001 \\
& Learning rate scheduler & Cosine Annealing & Cosine Annealing \\
& Weight decay & 0.0001 & 0.0001 \\
& Fourier layers & 4 & 4 \\
& Spatial Fourier modes & 22 & 22 \\
& Width & 10 & 10 \\
\hline
\multirow{11}{*}{\textbf{PCNO-2D}} & Data normalization & [0,1] & [0,1] \\
& Training strategies & Markov & Markov \\
& Epochs & 100 & 100 \\
& Batch size & 10 & 10 \\
& Optimizer & Adam & Adam \\
& Learning rate & 0.001 & 0.001 \\
& Learning rate scheduler & Cosine Annealing & Cosine Annealing \\
& Weight decay & 0.0001 & 0.0001 \\
& Fourier layers & 4 & 4 \\
& Spatial Fourier modes & 22 & 22 \\
& Width & 20 & 20 \\
\hline
\end{tabular}%
\end{table}
\begin{table}[ht!]
\centering
\begin{tabular}{cccc}
\hline
\textbf{Methods} & \textbf{Parameters} & \textbf{Transformed variables} & \textbf{Atmospheric variables} \\
\hline
\multirow{12}{*}{\textbf{DiffPCNO}} & Data normalization & [0,1] & [0,1] \\
& Training strategies & Markov & Markov \\
& Epochs & 100 & 100 \\
& Batch size & 10 & 10 \\
& Optimizer & Adam & Adam \\
& Betas & (0.9, 0.995) & (0.9, 0.995) \\
& Learning rate & 0.0001 & 0.0001 \\
& Learning rate scheduler & LinearLR & LinearLR \\
& Start factor & 1.00E-05 & 1.00E-05 \\
& Total iters & 1000 & 1000 \\
& Network                            & Modern U-Net \\
& Hidden channels & 64 & 64 \\
\hline
\multirow{9}{*}{\textbf{U-Net-3D}} & Training strategies & One-shot & - \\
& Epochs & 500 & - \\
& Batch size & 10 & - \\
& Optimizer & Adam & - \\
& Learning rate & 0.001 & - \\
& Learning rate scheduler & Cosine Annealing & - \\
& Weight decay & 0.0001 & - \\
& Feature dimension of the first layer & 32 & - \\
& Time padding & 6 & - \\
\hline
\multirow{12}{*}{\textbf{FNO-3D}} & Training strategies & One-shot & - \\
& Epochs & 500 & - \\
& Batch size & 10 & - \\
& Optimizer & Adam & - \\
& Learning rate & 0.001 & - \\
& Learning rate scheduler & Cosine Annealing & - \\
& Weight decay & 0.0001 & - \\
& Fourier layers & 4 & - \\
& Spatial Fourier modes & 22 & - \\
& Time Fourier modes & 8 & - \\
& Width & 20 & - \\
& Time padding & 6 & - \\
\hline
\multirow{12}{*}{\textbf{G-FNO-3D}} & Training strategies & One-shot & - \\
& Epochs & 500 & - \\
& Batch size & 10 & - \\
& Optimizer & Adam & - \\
& Learning rate & 0.001 & - \\
& Learning rate scheduler & Cosine Annealing & - \\
& Weight decay & 0.0001 & - \\
& Fourier layers & 4 & - \\
& Spatial Fourier modes & 22 & - \\
& Time Fourier modes & 8 & - \\
& Width & 11 & - \\
& Time padding & 6 & - \\
\hline
\multirow{12}{*}{\textbf{PCNO-3D}}  & Training strategies & One-shot & - \\
& Epochs & 500 & - \\
& Batch size & 10 & - \\
& Optimizer & Adam & - \\
& Learning rate & 0.001 & - \\
& Learning rate scheduler & Cosine Annealing & - \\
& Weight decay & 0.0001 & - \\
& Fourier layers & 4 & - \\
& Spatial Fourier modes & 22 & - \\
& Time Fourier modes & 8 & - \\
& Width & 20 & - \\
& Time padding & 6 & - \\
\hline
\end{tabular}%
\end{table}

\begin{table}[htbp]
\centering
\caption{Runtime of the models and hydrodynamic method for flood inundation forecasting.}
\label{table06}
\begin{tabular}{cccccc}
\hline
\textbf{\makecell{Flood \\ events}}                   & \textbf{Method}                                             & \textbf{\makecell{Hydrodynamic \\method}} & \textbf{\makecell{FNO \\(training\\ $T$=24h)}} & \textbf{\makecell{PCNO\\ (training\\ $T$=24h)}} & \textbf{\makecell{DiffPCNO \\(training\\ $T$=24h)}} \\ \hline
\multirow{3}{*}{\textbf{\makecell{Pakistan \\flood}}}  & Training time (hours)                                        & -                                       & 75.12                         & 70.22                          & 369.95                             \\
                                          & \makecell{Prediction   time for two-day \\480m Pakistan flood (seconds)} & -                                       & 25.82                         & 42.01                          & 1513.94                            \\
                                          & \makecell{Simulation   time for two-day \\480m Pakistan flood (hours)}  & 24                                      & -                             & -                              & -                                  \\ \hline
\multirow{3}{*}{\textbf{\makecell{Australia \\flood}}} & Training time (hours)                                        & -                                       & 50.27                         & 45.77                          & 295.84                             \\
                                          & \makecell{Prediction   time  for one-day \\30m Australia flood (seconds) }& -                                       & 20.94                         & 20.92                          & 1023.90                                   \\
                                          & \makecell{Simulation   time  for one-day \\30m Australia flood (hours)}    & 4.8                                     & -                             & -                              & -                                  \\ \hline
\end{tabular}
\end{table}

\begin{table}[htbp]
\centering
\caption{Runtime of the models for atmospheric modeling.}
\label{table3}
\begin{tabular}{cccccc}
\hline
\textbf{Method}                   & \textbf{U-Net-2D} & \textbf{FNO-2D}  & \textbf{G-FNO-2D}    & \textbf{PCNO-2D}     & \textbf{DiffPCNO} \\
\hline
Training time(hours)     & 3.71     & 3.57    & 8.72        & 7.07        & 21.20     \\
Prediction time(seconds) & 0.026 & 0.027 & 0.031 & 0.036 & 0.629 \\
\hline
\end{tabular}
\end{table}

\end{document}